\documentclass[12pt,twoside]{article}
\usepackage{latexsym}
\makeindex

\def\,{\mskip 3mu} \def\>{\mskip 4mu plus 2mu minus 4mu} \def\;{\mskip 5mu plus 5mu} \def\!{\mskip-3mu}
\def\dispmuskip{\thinmuskip= 3mu plus 0mu minus 2mu \medmuskip=  4mu plus 2mu minus 2mu \thickmuskip=5mu plus 5mu minus 2mu}
\def\textmuskip{\thinmuskip= 0mu                    \medmuskip=  1mu plus 1mu minus 1mu \thickmuskip=2mu plus 3mu minus 1mu}
\textmuskip
\def\beq{\dispmuskip\begin{equation}}    \def\eeq{\end{equation}\textmuskip}
\def\beqn{\dispmuskip\begin{displaymath}}\def\eeqn{\end{displaymath}\textmuskip}
\def\bqa{\dispmuskip\begin{eqnarray}}    \def\eqa{\end{eqnarray}\textmuskip}
\def\bqan{\dispmuskip\begin{eqnarray*}}  \def\eqan{\end{eqnarray*}\textmuskip}

\newenvironment{keywords}{\centerline{\bf\small
Keywords}\vspace{-2ex}\begin{quote}\small}{\par\end{quote}\vskip 1ex}
\newtheorem{theorem}{Theorem}
\newtheorem{corollary}[theorem]{Corollary}
\newtheorem{lemma}[theorem]{Lemma}
\newtheorem{definition}[theorem]{Definition}
\def\lessgtr{{\stackrel{\displaystyle<}>}}

\def\ftheorem#1#2#3{\begin{theorem}[#2]\label{#1} #3 \end{theorem} }

\def\fdefinition#1#2#3{\begin{definition}[#2]\label{#1} #3 \end{definition} }

\def\idx#1{\index{#1}#1} 
\def\indxs#1#2{\index{#1!#2}\index{#2!#1}} 
\def\tcite{\cite}
\def\citey{\cite}
\def\citein#1{in \cite{#1}}
\def\citealt{\cite}
\def\paradot#1{\vspace{1ex}\noindent{\bf #1.}}
\def\paranodot#1{\vspace{1ex}\noindent{\bf #1}}
\def\toinfty#1{\stackrel{#1\to\infty}{\longrightarrow}}
\def\nq{\hspace{-1em}}
\def\qed{\hspace*{\fill}$\Box\quad$}
\def\odt{{\textstyle{1\over 2}}}
\def\eps{\varepsilon}                   
\def\epstr{\epsilon}                    
\def\v#1{{\bf #1}}
\def\approxleq{\mbox{\raisebox{-0.8ex}{$\stackrel{\displaystyle<}\sim$}}} 
\def\approxgeq{\mbox{\raisebox{-0.8ex}{$\stackrel{\displaystyle>}\sim$}}} 
\def\M{{\cal M}}
\def\X{{\cal X}}
\def\Y{{\cal Y}}
\def\F{{\cal F}}
\def\E{{\bf E}}
\def\P{{\bf P}}
\def\Km{K\!m}
\def\Set#1{\if#1Q{I\!\!\!#1}\else\if#1Z{Z\!\!\!Z}\else{I\!\!#1}\fi\fi}
\def\qmbox#1{{\quad\mbox{#1}\quad}}

\begin{document}
\begin{titlepage}

\title{\vskip -10mm \normalsize\sc Technical Report IDSIA-02-02 \hfill February 2002 -- January 2003
  \vskip 2mm\bf\LARGE\hrule height5pt \vskip 5mm
  \Large\sc Optimality of Universal Bayesian Sequence \\
          Prediction for General Loss and Alphabet
  \vskip 6mm \hrule height2pt \vskip 4mm}
\author{{\bf Marcus Hutter}\\[3mm]
\normalsize IDSIA, Galleria 2, CH-6928\ Manno-Lugano, Switzerland\\
\normalsize marcus@idsia.ch \hspace{8.5ex} http://www.idsia.ch/$^{_{_\sim}}\!$marcus}
\date{}
\maketitle

\begin{abstract}
\noindent Various optimality properties of universal sequence
predictors based on Bayes-mixtures in general, and Solomonoff's
prediction scheme in particular, will be studied.
The probability of observing $x_t$ at time $t$, given past
observations $x_1...x_{t-1}$ can be computed with the chain rule
if the true generating distribution $\mu$ of the sequences
$x_1x_2x_3...$ is known. If $\mu$ is unknown, but known to belong
to a countable or continuous class $\M$ one can base ones
prediction on the Bayes-mixture $\xi$ defined as a
$w_\nu$-weighted sum or integral of distributions $\nu\in\M$. The
cumulative expected loss of the Bayes-optimal universal prediction
scheme based on $\xi$ is shown to be close to the loss of the
Bayes-optimal, but infeasible prediction scheme based on $\mu$. We
show that the bounds are tight and that no other predictor can
lead to significantly smaller bounds.
Furthermore, for various performance measures, we show
Pareto-optimality of $\xi$ and give an Occam's razor argument that
the choice $w_\nu\sim 2^{-K(\nu)}$ for the weights is optimal,
where $K(\nu)$ is the length of the shortest program describing
$\nu$.
The results are applied to games of chance, defined as a sequence
of bets, observations, and rewards.
The prediction schemes (and bounds) are compared to the popular
predictors based on expert advice.
Extensions to infinite alphabets, partial, delayed and
probabilistic prediction, classification, and more active systems
are briefly discussed.
\end{abstract}

\begin{keywords}%
  Bayesian sequence prediction; mixture distributions; Solomonoff
  induction; Kolmogorov complexity; learning; universal probability;
  tight loss and error bounds; Pareto-optimality; games of chance;
  classification.
\end{keywords}

\end{titlepage}

\parskip=0ex\tableofcontents\newpage

\section{Introduction}\label{secInt}
Many problems are of the induction type in which statements about
the future have to be made, based on past observations. What is
the probability of rain tomorrow, given the weather observations
of the last few days? Is the Dow Jones likely to rise tomorrow,
given the chart of the last years and possibly additional
newspaper information? Can we reasonably doubt that the sun will
rise tomorrow? Indeed, one definition of science is to predict the
future, where, as an intermediate step, one tries to understand
the past by developing theories and finally to use the prediction
as the basis for some decision. Most induction problems can be
studied in the Bayesian framework. The probability of observing
$x_t$ at time $t$, given the observations $x_1...x_{t-1}$ can be
computed with the chain rule, if we know the true probability
distribution, which generates the observed sequence
$x_1x_2x_3...$. The problem is that in many cases we do not even
have a reasonable guess of the true distribution $\mu$. What is
the true probability of weather sequences, stock charts, or
sunrises?

\index{sequence prediction!universal}
In order to overcome the problem of the unknown true distribution,
one can define a mixture distribution $\xi$ as a weighted
sum or integral over distributions $\nu\in\cal M$, where $\cal M$
is any discrete or continuous (hypothesis) set including $\mu$.
$\M$ is assumed to be known and to contain the true distribution,
i.e.\ $\mu\in\M$.
Since the probability $\xi$ can be shown to converge rapidly to
the true probability $\mu$ in a conditional sense, making
decisions based on $\xi$ is often nearly as good as the infeasible
optimal decision based on the unknown $\mu$ \cite{Merhav:98}.
Solomonoff \citey{Solomonoff:64} had the idea to define a
universal mixture as a weighted average over deterministic
programs. Lower weights were assigned to longer programs. He
unified Epicurus' principle of multiple explanations and Occam's
razor [simplicity] principle into one formal theory (See
\citealt{Li:97} for this interpretation of
\citealt{Solomonoff:64}). Inspired by Solomonoff's idea, Levin
\citey{Zvonkin:70} defined the closely related universal prior
$\xi_U$ as a weighted average over {\em all} semi-computable
probability distributions. If the environment possesses some
effective structure at all, Solomonoff-Levin's posterior ``finds''
this structure \cite{Solomonoff:78}, and allows for a good
prediction. In a sense, this solves the induction problem in a
universal way, i.e.\ without making problem specific assumptions.

\paranodot{Section \ref{secSetup}} explains notation and defines the
{\em universal or mixture distribution} $\xi$ as the
$w_\nu$-weighted sum of probability distributions $\nu$ of a set
$\M$, which includes the true distribution $\mu$. No structural
assumptions are made on the $\nu$. $\xi$ multiplicatively
dominates all $\nu \in \M$, and the relative entropy between $\mu$
and $\xi$ is bounded by $\ln{w_\mu^{-1}}$. Convergence of $\xi$ to
$\mu$ in a mean squared sense is shown in Theorem \ref{thConv}.
The representation of the universal posterior distribution and the
case $\mu\not\in\M$ are briefly discussed. Various standard sets
$\M$ of probability measures are discussed, including computable,
enumerable, cumulatively enumerable, approximable, finite-state, and
Markov (semi)measures.

\paranodot{Section \ref{secErr}} is essentially a generalization of the
deterministic error bounds found \citein{Hutter:99errbnd} from the
binary alphabet to a general finite alphabet $\X$. Theorem
\ref{thErrBnd} bounds $E^{\Theta_\xi} - E^{\Theta_\mu}$ by
$O(\sqrt{E^{\Theta_\mu}})$, where $E^{\Theta_\xi}$ is the expected
number of errors made by the optimal universal predictor
$\Theta_\xi$, and $E^{\Theta_\mu}$ is the expected number of
errors made by the optimal informed prediction scheme
$\Theta_\mu$. The non-binary setting cannot be reduced to the
binary case! One might think of a binary coding of the symbols
$x_t\in\X$ in the sequence $x_1x_2...$. But this makes it
necessary to predict a block of bits $x_t$, before one receives
the true block of bits $x_t$, which differs from the bit by bit
prediction scheme considered
\citein{Solomonoff:78,Hutter:99errbnd}.
The framework generalizes to the case where an action $y_t \in \Y$
results in a loss $\ell_{x_t y_t}$ if $x_t$ is the next symbol of
the sequence. Optimal universal $\Lambda_\xi$ and optimal informed
$\Lambda_\mu$ prediction schemes are defined for this case, and
loss bounds similar to the error bounds of the last section are
stated. No assumptions on $\ell$ have to be made, besides
boundedness.

\paranodot{Section \ref{secGames}} applies
the loss bounds to games of chance, defined as a sequence of bets,
observations, and rewards. The average profit $\bar
p_n^{\Lambda_\xi}$ achieved by the $\Lambda_\xi$ scheme rapidly
converges to the best possible average profit $\bar
p_n^{\Lambda_\mu}$ achieved by the $\Lambda_\mu$ scheme ($\bar
p_n^{\Lambda_\xi} - \bar p_n^{\Lambda_\mu} = O(n^{-1/2})$). If
there is a profitable scheme at all ($\bar p_n^{\Lambda_\mu} >
\eps > 0$), asymptotically the universal $\Lambda_\xi$ scheme will
also become profitable. Theorem
\ref{thWin} bounds the time needed to reach the winning zone. It
is proportional to the relative entropy of $\mu$ and $\xi$ with a
factor depending on the profit range and on $\bar
p_n^{\Lambda_\mu}$. An attempt is made to give an information
theoretic interpretation of the result.

\paranodot{Section \ref{secLower}} discusses the quality of the
universal predictor and the bounds. We show that
there are $\M$ and $\mu\in\M$ and weights $w_\nu$ such that the
derived error bounds are tight. This shows that the error bounds
cannot be improved in general. We also show Pareto-optimality of
$\xi$ in the sense that there is no other predictor which performs
at least as well in all environments $\nu\in\M$ and strictly
better in at least one. Optimal predictors can always be based on
mixture distributions $\xi$. This still leaves open how to choose
the weights. We give an Occam's razor argument that the choice
$w_\nu=2^{-K(\nu)}$, where $K(\nu)$ is the length of the shortest
program describing $\nu$ is optimal.

\paranodot{Section \ref{secMisc}} generalizes the setup to continuous
probability classes $\M = \{\mu_\theta\}$ consisting of continuously
parameterized distributions $\mu_\theta$ with parameter
$\theta \in \Set{R}^d$. Under certain smoothness and regularity
conditions a bound for the relative entropy between $\mu$ and
$\xi$, which is central for all presented results, can still be
derived. The bound depends on the Fisher information of $\mu$ and
grows only logarithmically with $n$, the intuitive reason being
the necessity to describe $\theta$ to an accuracy $O(n^{-1/2})$.
Furthermore, two ways of using the prediction schemes for partial
sequence prediction, where not every symbol needs to be predicted,
are described. Performing and predicting a sequence of independent
experiments and online learning of classification tasks are
special cases.
We also compare the universal prediction scheme studied here to
the popular predictors based on expert advice (PEA)
\cite{Littlestone:89,Vovk:92,Littlestone:94,Cesa:97,Haussler:98,Kivinen:99}.
Although the algorithms, the settings, and the proofs are quite
different, the PEA bounds and our error bound have the same
structure.
Finally, we outline possible extensions of the presented theory
and results, including infinite alphabets, delayed and
probabilistic prediction, active systems influencing the
environment, learning aspects, and a unification with PEA.

\paranodot{Section \ref{secSPConc}} summarizes the results.

There are good introductions and surveys of Solomonoff sequence
prediction \cite{Li:92b,Li:97}, inductive inference in general
\cite{Angluin:83,Solomonoff:97,Merhav:98}, reasoning under
uncertainty \cite{Gruenwald:98}, and competitive online statistics
\cite{Vovk:99}, with interesting relations to this work. See
Section \ref{secWM} for some more details.

\section{Setup and Convergence}\label{secSetup}

In this section we show that the mixture $\xi$ converges rapidly to
the true distribution $\mu$.
After defining basic notation in Section \ref{secRanSeq},
we introduce in Section \ref{secUPPD} the {\em universal or
mixture distribution} $\xi$ as the $w_\nu$-weighted sum of
probability distributions $\nu$ of a set $\M$, which includes the
true distribution $\mu$. No structural assumptions are made on the
$\nu$. $\xi$ multiplicatively dominates all $\nu\in\M$.
A posterior representation of $\xi$ with incremental weight update
is presented in Section \ref{secUPost}.
In Section \ref{secConv} we show that the relative entropy between $\mu$
and $\xi$ is bounded by $\ln{w_\mu^{-1}}$ and that $\xi$ converges to
$\mu$ in a mean squared sense.
The
case $\mu\not\in\M$ is briefly discussed in Section \ref{subsecnotM}.
The section concludes with Section \ref{secPCM}, which
discusses various standard sets $\M$ of probability measures,
including computable, enumerable, cumulatively enumerable,
approximable, finite-state, and Markov (semi)measures.

\subsection{Random Sequences}\label{secRanSeq}\indxs{sequence}{random}
We denote strings over a finite alphabet $\X$ by $x_1x_2...x_n$
with $x_t\in\X$ and $t,n,N\in\Set N$ and $N=|\X|$. We further use
the abbreviations $\epstr$\index{string!empty} for the empty
string, $x_{t:n}:=x_tx_{t+1}...x_{n-1}x_n$ for $t\leq n$ and
$\epstr$ for $t>n$, and $x_{<t}:=x_1... x_{t-1}$. We use Greek
letters for probability distributions (or measures). Let
$\rho(x_1...x_n)$ be the probability that an (infinite) sequence
starts with $x_1...x_n$:
\beqn\label{propsp}
  \sum_{x_{1:n}\in\X^n}\rho(x_{1:n})=1,\quad
  \sum_{x_t\in\X}\rho(x_{1:t}) =
  \rho(x_{<t})
  ,\quad
  \rho(\epstr)=1.
\eeqn
We also need conditional probabilities derived from the chain
rule:
\bqan\label{bayes}
  \rho(x_t|x_{<t}) &=&
  \rho(x_{1:t})/\rho(x_{<t}),\quad
  \\[4mm]\label{bayesn}
  \rho(x_1...x_n) &=&
  \rho(x_1) \!\cdot\!
  \rho(x_2|x_1)
  \!\cdot\!...\!\cdot\!
  \rho(x_n|x_1...x_{n-1}).
\eqan
The first equation states that the probability that a string
$x_1...x_{t-1}$ is followed by $x_t$ is equal to the probability
that a string starts with $x_1...x_t$ divided by the probability
that a string starts with $x_1...x_{t-1}$. For convenience we
define $\rho(x_t|x_{<t}) = 0$ if $\rho(x_{<t}) = 0$.
The second equation is the first, applied $n$ times. Whereas
$\rho$ might be any probability distribution, $\mu$ denotes the
true (unknown) generating distribution of the sequences. We
denote probabilities by $\P$, expectations by $\E$ and further abbreviate
\beqn
\E_t[..]:=\sum_{x_t\in\X}\mu(x_t|x_{<t})[..],\qquad
\E_{1:n}[..]:=\nq\sum_{x_{1:n}\in\X^n}\!\!\!\mu(x_{1:n})[..],\qquad
\E_{<t}[..]:=\nq\sum_{x_{<t}\in\X^{t-1}}\nq\mu(x_{<t})[..].
\eeqn
Probabilities $\P$ and expectations $\E$ are {\em always} w.r.t.\
the true distribution $\mu$. $\E_{1:n} = \E_{<n}\E_n$ by the chain
rule and $\E[...]=\E_{<t}[...]$ if the argument is independent of
$x_{t:\infty}$, and so on. We abbreviate ``with $\mu$-probability
1'' by w.$\mu$.p.1.%
\index{convergence!in mean sum}
\index{convergence!in the mean}
\index{convergence!in probability}
\index{convergence!with probability 1}%
We say that $z_t$ converges to $z_*$ {\em in mean sum} (i.m.s.) if
$c:=\sum_{t=1}^\infty\E[(z_t-z_*)^2]<\infty$. One can show that
convergence in mean sum implies convergence with probability 1.%
\footnote{Convergence in the mean, i.e.\
$\E[(z_t-z_*)^2]\toinfty{t} 0$, only implies convergence in
probability, which is weaker than convergence with probability 1.}
Convergence i.m.s.\ is very strong: it provides a ``rate'' of
convergence in the sense that the expected number of times $t$ in
which $z_t$ deviates more than $\eps$ from $z_*$ is finite and
bounded by $c/\eps^2$ and the
probability that the number of $\eps$-deviations exceeds
$c\over\eps^2\delta$ is smaller than $\delta$.

\subsection{Universal Prior Probability Distribution}\label{secUPPD}
\indxs{universal}{probability distribution}\indxs{prior}{probability}
Every inductive inference problem can be brought into the
following form: Given a string $x_{<t}$, take a guess at its
continuation $x_t$. We will assume that the strings which have to
be continued are drawn from a probability\footnote{This includes
deterministic environments, in which case the probability
distribution $\mu$ is $1$ for some sequence $x_{1:\infty}$ and $0$
for all others. We call probability distributions of this kind
{\em deterministic}.} distribution $\mu$.
\index{probability distribution!true}%
\index{probability distribution!generating}%
\index{probability distribution!unknown}%
The maximal prior information a prediction algorithm can possess
is the exact knowledge of $\mu$, but in many cases (like for the
probability of sun tomorrow) the true generating distribution is
not known. Instead, the prediction is based on a guess $\rho$ of
$\mu$. We expect that a predictor based on $\rho$ performs well,
if $\rho$ is close to $\mu$ or converges, in a sense, to $\mu$.
Let $\M := \{\nu_1,\nu_2,...\}$ be a countable set of candidate
probability distributions on strings. Results are generalized to
continuous sets $\M$ in Section \ref{secCPC}. We define a weighted
average on $\M$
\beq\label{xidefsp}
  \xi(x_{1:n}) \;:=\;
  \sum_{\nu\in\M}w_\nu\!\cdot\!\nu(x_{1:n}),\quad
  \sum_{\nu\in\M}w_\nu=1,\quad w_\nu>0.
\eeq
It is easy to see that $\xi$ is a probability distribution as the
\idx{weights} $w_\nu$ are positive and normalized to 1 and the
$\nu \in \M$ are probabilities.\footnote{The weight $w_\nu$ may be
interpreted as the initial degree of belief in $\nu$ and
$\xi(x_1...x_n)$ as the degree of belief in $x_1...x_n$. If the
existence of true randomness is rejected on philosophical grounds
one may consider $\M$ containing only deterministic environments.
$\xi$ still represents belief
probabilities\indxs{belief}{probability}.} For a finite $\M$ a
possible choice for the $w$ is to give all $\nu$ equal weight
($w_\nu={1\over|\M|}$). We call $\xi$ universal relative to $\M$,
as it multiplicatively dominates all distributions in $\M$
\index{multiplicative!domination}
\beq\label{unixi}
  \xi(x_{1:n}) \;\geq\;
  w_\nu\!\cdot\!\nu(x_{1:n}) \quad\mbox{for all}\quad
  \nu\in\M.
\eeq
In the following, we assume that $\M$ is known and contains the
true distribution, i.e.\ $\mu\in\M$. If $\M$ is chosen
sufficiently large, then $\mu\in\M$ is not a serious
constraint.

\subsection{Universal Posterior Probability Distribution}\label{secUPost}
\index{probability distribution!posterior}\index{posterior probability}
\index{probability distribution!conditional}
\index{weights!time dependent}\index{weights!posterior}
All prediction schemes in this work are based on the conditional
probabilities $\rho(x_t|x_{<t})$. It is possible to express
also the conditional probability $\xi(x_t|x_{<t})$ as a
weighted average over the conditional $\nu(x_t|x_{<t})$, but
now with time dependent weights:
\beq\label{xirel}
  \xi(x_t|x_{<t})=\sum_{\nu\in\M}w_\nu(x_{<t})\nu(  x_t|x_{<t}), \quad
  w_\nu(x_{1:t}):=w_\nu(x_{<t}){\nu(x_t|x_{<t}) \over
  \xi(x_t|x_{<t})}, \quad
  w_\nu(\epstr):=w_\nu.
\eeq
The denominator just ensures correct normalization $\sum_\nu
w_\nu(x_{1:t}) = 1$. By induction and the chain rule we see that
$
  w_\nu(x_{<t}) = w_\nu\nu(x_{<t})/\xi(x_{<t})
$.
Inserting this into $\sum_\nu w_\nu(x_{<t})\nu(x_t|x_{<t})$
using (\ref{xidefsp}) gives $\xi(x_t|x_{<t})$, which proves
the equivalence of (\ref{xidefsp}) and (\ref{xirel}). The
expressions (\ref{xirel}) can be used to give an intuitive, but
non-rigorous, argument why $\xi(x_t|x_{<t})$ converges to
$\mu(x_t|x_{<t})$:
The weight $w_\nu$ of $\nu$ in $\xi$ increases/decreases if $\nu$
assigns a high/low probability to the new symbol $x_t$, given
$x_{<t}$. For a $\mu$-random sequence $x_{1:t}$, $\mu(x_{1:t})\gg
\nu(x_{1:t})$ if $\nu$ (significantly) differs from $\mu$. We
expect the total weight for all $\nu$ consistent with $\mu$ to
converge to 1, and all other weights to converge to 0 for
$t\to\infty$. Therefore we expect $\xi(x_t|x_{<t})$ to converge to
$\mu(x_t|x_{<t})$ for $\mu$-random strings $x_{1:\infty}$.

Expressions (\ref{xirel}) seem to be more suitable than
(\ref{xidefsp}) for studying convergence and loss bounds of the
universal predictor $\xi$, but it will turn out that (\ref{unixi})
is all we need, with the sole exception in the proof of Theorem
\ref{thPareto}. Probably (\ref{xirel}) is useful when one tries to
understand the \idx{learning} aspect in $\xi$.

\subsection{Convergence of $\xi$ to $\mu$}\label{secConv}
We use the relative entropy and the squared Euclidian/absolute
distance to measure the instantaneous and total distances
between $\mu$ and $\xi$:
\beq\label{entropy}\label{dtmuxi}\label{DNmuxi}
  d_t(x_{<t}) \;:=\;
  \E_t\ln{\mu(x_t|x_{<t}) \over \xi(
  x_t|x_{<t})}
  ,\qquad
  D_n \;:=\; \sum_{t=1}^n \E_{<t} d_t(x_{<t}) \;=\;
  \E_{1:n} \ln{\mu(x_{1:n}) \over \xi(x_{1:n})}
\eeq\vspace{-2ex}
\beq\label{eukdistxi1}\label{stmuxi}
  s_t(x_{<t}) \;:=\; \sum_{x_t}
  \Big(\mu(x_t|x_{<t})-\xi(x_t|x_{<t})\Big)^2
  ,\qquad\qquad
  S_n \;:=\; \sum_{t=1}^n\E_{<t}s_t(x_{<t})
\eeq
\beq\label{eukdistxi2}
  \quad a_t(x_{<t}) \;:=\; \sum_{x_t}
  \Big|\mu(x_t|x_{<t})-\xi(x_t|x_{<t})\Big|
  ,\qquad\qquad\quad
  V_n \;:=\; {1\over 2}\sum_{t=1}^n\E_{<t}a^2_t(x_{<t})
\eeq
One can show that $s_t \leq \odt a^2_t \leq d_t$
\cite[Sec.3.2]{Hutter:01alpha} \cite[Lem.12.6.1]{Cover:91}, hence
$S_n\leq V_n\leq D_n$ (for binary alphabet, $s_t=\odt a^2_t$,
hence $S_n=V_n$). So bounds in terms of $S_n$ are tightest, while
the (implied) looser bounds in terms of $V_n$ as a referee pointed
out have an advantage in case of continuous alphabets (not
considered here) to be reparametrization-invariant. The weakening
to $D_n$ is used, since $D_n$ can easily be bounded in terms of
the weight $w_\mu$.

\ftheorem{thConv}{Convergence}{
Let there be sequences $x_1x_2...$ over a finite alphabet
$\X$ drawn with probability $\mu(x_{1:n})$ for the first
$n$ symbols. The universal conditional probability $\xi(x_t|x_{<t})$
of the next symbol $x_t$ given $x_{<t}$ 
is related to the true conditional probability
$\mu(x_t|x_{<t})$ in the following way:
\beqn
        \sum_{t=1}^n\E_{<t}\sum_{x_t}
        \Big(\mu(x_t|x_{<t})-\xi(x_t|x_{<t})\Big)^2
        \;\equiv\; S_n \;\leq\; V_n \;\leq\;
        D_n \;\leq\; \ln w_\mu^{-1} =: b_\mu \;<\; \infty
\eeqn
where $d_t$ and $D_n$ are the relative entropies (\ref{dtmuxi}),
and $w_\mu$ is the weight (\ref{xidefsp}) of
$\mu$ in $\xi$.
}

A proof for binary alphabet can be found in
\tcite{Solomonoff:78,Li:97} and for a general finite alphabet in
\tcite{Hutter:01alpha}. The finiteness of $S_\infty$ implies
$\xi(x'_t|x_{<t}) - \mu(x'_t|x_{<t}) \to 0$ for $t\to\infty$
i.m.s., and hence w.$\mu$.p.1 for any $x'_t$. There are other
convergence results, most notably $\xi(
x_t|x_{<t})/\mu(x_t|x_{<t}) \to 1$ for $t\to\infty$ w.$\mu$.p.1
\cite{Li:97,Hutter:02spupper}. These convergence results motivate
the belief that predictions based on (the known) $\xi$ are
asymptotically as good as predictions based on (the unknown) $\mu$
with rapid convergence.

\subsection{The Case where $\mu \not\in \M$}\label{subsecnotM}
In the following we discuss two cases, where $\mu \not\in \M$,
but most parts of this work still apply.
Actually all theorems remain valid for $\mu$ being a finite linear
combination $\mu(x_{1:n})=\sum_{\nu\in\cal L}v_{\nu} \nu(x_{1:n})$
of $\nu$'s in ${\cal L} \subseteq \M$. Dominance $\xi(x_{1:n})\geq
w_\mu\cdot\mu(x_{1:n})$ is still ensured with
$w_\mu:=\min_{\nu\in\cal L}{w_\nu\over
v_{\nu}}\geq\min_{\nu\in\cal L}w_\nu$.
More generally, if $\mu$ is an infinite linear combination,
dominance is still ensured if $w_\nu$ itself dominates
$v_{\nu}$ in the sense that $w_\nu \geq \alpha v_{\nu}$
for some $\alpha > 0$ (then $w_\mu \geq \alpha$).

\index{probability!nearby distribution}\index{probability distribution!nearby}
Another possibly interesting situation is when the true generating
distribution $\mu \not\in \M$, but a ``nearby'' distribution
$\hat\mu$ with weight $w_{\hat\mu}$ is in $\M$. If we measure
the distance of $\hat\mu$ to $\mu$ with the Kullback-Leibler
divergence
$D_n(\mu||\hat\mu):=\sum_{x_{1:n}}\mu(x_{1:n})\ln{\mu(x_{1:n})\over\hat\mu(x_{1:n})}$
and assume that it is bounded by a constant $c$, then
\beqn
  D_n \;=\; \E_{1:n} \ln{\mu(x_{1:n})\over\xi(x_{1:n})} \;=\;
  \E_{1:n} \ln{\hat\mu(x_{1:n})\over\xi(x_{1:n})} +
  \E_{1:n} \ln{\mu(x_{1:n})\over\hat\mu(x_{1:n})} \;\leq\;
  \ln w_{\hat\mu}^{-1} + c.
\eeqn
So $D_n\!\leq\ln w_\mu^{-1}$ remains valid if we define
$w_\mu := w_{\hat\mu} \cdot e^{-c}$.

\subsection{Probability Classes $\M$}\label{secPCM}
\index{probability!classes}
\indxs{probability class}{discrete}
\indxs{probability class}{countable}
In the following we describe some well-known and some less known
probability classes $\M$. This relates our setting to other
works in this area, embeds it into the historical context,
illustrates the type of classes we have in mind, and discusses
computational issues.

\index{probability distribution!computable}
We get a rather wide class $\M$ if we include {\em all}
(semi)computable probability distributions in $\M$. In this case,
the assumption $\mu\in\M$ is very weak, as it only assumes
that the strings are drawn from {\em any (semi)computable} distribution;
and all valid physical theories (and, hence, all environments)
{\em are} computable to arbitrary precision (in a probabilistic sense).

\index{probability distribution!simple}
We will see that it is favorable to assign high weights $w_\nu$ to the
$\nu$. Simplicity should be favored over complexity, according to
Occam's razor. In our context this means that a high weight should
be assigned to simple $\nu$. The prefix Kolmogorov complexity
$K(\nu)$ is a universal complexity measure
\cite{Kolmogorov:65,Zvonkin:70,Li:97}. It is defined as the length of
the shortest self-delimiting program (on a universal Turing
machine) computing $\nu(x_{1:n})$ given $x_{1:n}$.
If we define
\beqn
  w_\nu \;:=\; 2^{-K(\nu)} \qquad
\eeqn
then distributions which can be calculated by short programs,
have high weights. The relative entropy is bounded by the
Kolmogorov complexity of $\mu$ in this case ($D_n \leq K(\mu)
\cdot \ln 2$).
\index{probability distribution!Solomonoff}%
\indxs{probability distribution}{universal}%
\indxs{Solomonoff}{semi-measure}%
\indxs{enumerable}{semi-measure}%
Levin's universal semi-measure $\xi_U$
is obtained if we take $\M=\M_U$
to be the (multi)set enumerated by a Turing machine which
enumerates all enumerable semi-measures
\cite{Zvonkin:70,Li:97}.
Recently, $\M$ has been further enlarged to include all
cumulatively enumerable\indxs{cumulatively
enumerable}{semi-measure} semi-measures \cite{Schmidhuber:02gtm}.
In the enumerable and cumulatively enumerable cases, $\xi$ is not
finitely computable, but can still be approximated to arbitrary
but not pre-specifiable precision.%
\indxs{approximable}{probability distribution} If we consider {\em
all} approximable (i.e.\ asymptotically computable) distributions,
then the universal distribution $\xi$, although still well
defined, is not even approximable \cite{Hutter:03unipriors}. An
interesting and quickly
approximable\index{prior!Speed}\index{Speed prior} distribution is
the Speed prior $S$ defined \citein{Schmidhuber:02speed}. It is
related to Levin complexity and Levin search
\cite{Levin:73search,Levin:84}, but it is unclear for now, which
distributions are dominated by $S$. If one considers only
\idx{finite-state automata} instead of general Turing machines,
$\xi$ is related to the quickly computable, universal finite-state
prediction scheme of Feder et al.\ \citey{Feder:92}, which itself
is related to the famous Lempel-Ziv data compression
algorithm\index{Lempel-Ziv
compression}\index{compression!Lempel-Ziv}. If one has extra
knowledge on the source generating the sequence, one might further
reduce $\M$ and increase $w$. A detailed analysis of these and
other specific classes $\M$ will be given elsewhere. Note that
$\xi\in\M$ in the enumerable and cumulatively enumerable case, but
$\xi\not\in\M$ in the computable, approximable and finite-state
case. If $\xi$ is itself in $\M$, it is called a universal element
of $\M$ \cite{Li:97}. As we do not need this property here, $\M$
may be {\em any} countable set of distributions. In the following
sections we consider generic $\M$ and $w$.

We have discussed various discrete classes $\M$, which are
sufficient from a constructive or computational point of view. On
the other hand, it is convenient to also allow for continuous
classes $\M$. For instance, the class of {\em all} Bernoulli
processes with parameter $\theta\in[0,1]$ and uniform prior
$w_\theta\equiv 1$ is much easier to deal with than computable
$\theta$ only, with prior $w_\theta=2^{-K(\theta)}$. Other
important continuous classes are the class of i.i.d.\ and Markov
processes. Continuous classes $\M$ are considered in more detail
in Section \ref{secCPC}. \index{probability distribution!generic
class}

\section{Error Bounds}\label{secErr}
\indxs{error}{bound}

In this section we prove error bounds for predictors based on
the mixture $\xi$.
Section \ref{secBOP} introduces the concept of Bayes-optimal
predictors $\Theta_\rho$, minimizing $\rho$-expected error.
In Section \ref{secTENE} we bound $E^{\Theta_\xi} -
E^{\Theta_\mu}$ by $O(\sqrt{E^{\Theta_\mu}})$, where
$E^{\Theta_\xi}$ is the expected number of errors made by the
optimal universal predictor $\Theta_\xi$, and $E^{\Theta_\mu}$ is
the expected number of errors made by the optimal informed
prediction scheme $\Theta_\mu$.
The proof is deferred to Section \ref{secPTErrBnd}.
In Section \ref{secLoss} we generalize the framework to the case
where an action $y_t \in \Y$ results in a loss $\ell_{x_t y_t}$ if
$x_t$ is the next symbol of the sequence. Optimal universal
$\Lambda_\xi$ and optimal informed $\Lambda_\mu$ prediction
schemes are defined for this case, and loss bounds similar to the
error bounds are presented. No assumptions on $\ell$ have to be
made, besides boundedness.

\subsection{Bayes-Optimal Predictors}\label{secBOP}
\indxs{Bayes-optimal}{prediction}
We start with a very simple measure: making a wrong prediction
counts as one error, making a correct prediction counts as no
error. In \cite{Hutter:99errbnd} error bounds have been proven for
the binary alphabet $\X=\{0,1\}$. The following generalization to
an arbitrary alphabet involves only minor additional
complications, but serves as an introduction to the more
complicated model with arbitrary loss function.
\indxs{minimize}{error}
Let $\Theta_\mu$ be the optimal prediction scheme when the strings
are drawn from the probability distribution $\mu$, i.e.\ the
probability of $x_t$ given $x_{<t}$ is $\mu(x_t|x_{<t})$, and
$\mu$ is known. $\Theta_\mu$ predicts (by definition)
$x_t^{\Theta_\mu}$ when observing $x_{<t}$. The prediction is
erroneous if the true $t^{th}$ symbol is not $x_t^{\Theta_\mu}$.
The probability of this event is $1-\mu(x_t^{\Theta_\mu}|x_{<t})$.
It is minimized if $x_t^{\Theta_\mu}$
maximizes $\mu(x_t^{\Theta_\mu}|x_{<t})$. More generally, let
$\Theta_\rho$ be a prediction scheme predicting
$x_t^{\Theta_\rho} := \arg\max_{x_t}\rho(x_t|x_{<t})$ for some
distribution $\rho$. Every deterministic predictor can be
interpreted as maximizing some distribution.

\subsection{Total Expected Numbers of Errors}\label{secTENE}
\indxs{error}{total}\indxs{error}{instantaneous}\indxs{error}{expected}
The $\mu$-probability of making a wrong prediction for the
$t^{th}$ symbol and the total $\mu$-expected number of errors in
the first $n$ predictions of predictor $\Theta_\rho$ are
\beq\label{rhoerr}
  e_t^{\Theta_\rho}(x_{<t}) \;:=\; 1-\mu(x_t^{\Theta_\rho}|x_{<t})
  \quad,\quad
  E_n^{\Theta_\rho} \;:=\; \sum_{t=1}^n \E_{<t}
  e_t^{\Theta_\rho}(x_{<t}).
\eeq
If $\mu$ is known, $\Theta_\mu$ is obviously the best
prediction scheme in the sense of making the least number of expected
errors
\beq\label{Emuopt}
  E_n^{\Theta_\mu} \;\leq\;E_n^{\Theta_\rho} \qmbox{for any}
  \Theta_\rho,
\eeq
since
\beqn
  e_t^{\Theta_\mu}(x_{<t}) \;=\;
  1\!-\!\mu(x_t^{\Theta_\mu}|x_{<t}) \;=\;
  \min_{x_t}\{1\!-\!\mu(x_t|x_{<t})\} \;\leq\;
  1\!-\!\mu(x_t^{\Theta_\rho}|x_{<t}) \;=\;
  e_t^{\Theta_\rho}(x_{<t})
\eeqn
\indxs{universal}{prediction}%
for any $\rho$. Of special interest is the universal predictor
$\Theta_\xi$. As $\xi$ converges to $\mu$ the prediction of
$\Theta_\xi$ might converge to the prediction of the optimal
$\Theta_\mu$. Hence, $\Theta_\xi$ may not make many more errors
than $\Theta_\mu$ and, hence, any other predictor $\Theta_\rho$.
Note that $x_t^{\Theta_\rho}$ is a discontinuous function of
$\rho$ and $x_t^{\Theta_\xi}\to x_t^{\Theta_\mu}$ cannot be proven
from $\xi\to\mu$. Indeed, this problem occurs in related
prediction schemes, where the predictor has to be regularized so
that it is continuous \cite{Feder:92}.
\index{regularization}Fortunately this is not necessary here. We
prove the following error bound.

\ftheorem{thErrBnd}{Error Bound}{
\indxs{error}{bound}\indxs{informed}{prediction}
Let there be sequences $x_1x_2...$ over a finite alphabet
$\X$ drawn with probability $\mu(x_{1:n})$ for the first
$n$ symbols. The $\Theta_\rho$-system predicts by definition
$x_t^{\Theta_\rho} \in \X$ from $x_{<t}$, where
$x_t^{\Theta_\rho}$ maximizes $\rho(x_t|x_{<t})$. $\Theta_\xi$
is the universal prediction scheme based on the universal prior
$\xi$. $\Theta_\mu$ is the optimal informed prediction scheme.
The total $\mu$-expected number of prediction errors $E_n^{\Theta_\xi}$ and
$E_n^{\Theta_\mu}$ of $\Theta_\xi$ and $\Theta_\mu$ as defined in
(\ref{rhoerr}) are bounded in the following way
\beqn\label{th2}
  0 \leq E_n^{\Theta_\xi}\!\!-\!E_n^{\Theta_\mu} \leq
  \sqrt{2Q_n S_n} \leq
  \sqrt{2(E_n^{\Theta_\xi}\!\!+\! E_n^{\Theta_\mu})S_n} \leq
  S_n\!+\!\sqrt{4E_n^{\Theta_\mu}S_n\!+\!S_n^2} \leq
  2S_n\!+\!2\sqrt{E_n^{\Theta_\mu}S_n}
\eeqn
where $Q_n=\sum_{t=1}^n\E_{<t}q_t$ (with
$q_t(x_{<t}):=1-\delta_{x_t^{\Theta_\xi}x_t^{\Theta_\mu}}$) is the
expected number of non-optimal predictions made by $\Theta_\xi$
and $S_n \leq V_n \leq D_n \leq \ln{w_\mu^{-1}}$, where $S_n$ is
the squared Euclidian distance (\ref{stmuxi}), $V_n$ half of the
squared absolute distance (\ref{eukdistxi2}), $D_n$ the relative
entropy (\ref{entropy}), and $w_\mu$ the weight (\ref{xidefsp})
of $\mu$ in $\xi$.
}

\noindent The first two bounds have a nice structure, but the r.h.s.\
actually depends on $\Theta_\xi$, so they are not
particularly useful, but these are the major bounds we will prove,
the others follow easily. In Section \ref{secLower} we show
that the third bound is optimal. The last bound, which we discuss
in the following, has the same asymptotics as the third bound.
Note that the bounds hold for any (semi)measure $\xi$; only
$D_n\leq\ln_\mu w^{-1}$ depends on $\xi$ dominating $\mu$ with
domination constant $w_\mu$.

\indxs{finite}{error} First, we observe that Theorem
\ref{thErrBnd} implies that the number of errors
$E_\infty^{\Theta_\xi}$ of the universal $\Theta_\xi$ predictor is
finite if the number of errors $E_\infty^{\Theta_\mu}$ of the
informed $\Theta_\mu$ predictor is finite. In particular, this is
the case for deterministic $\mu$, as $E_n^{\Theta_\mu} \equiv 0$
in this case\footnote{Remember that we named a probability
distribution {\em deterministic} if it is 1 for exactly one
sequence and 0 for all others.}, i.e.\ $\Theta_\xi$ makes only a
finite number of errors on deterministic environments.
This can also be proven by elementary means. Assume $x_1x_2...$ is
the sequence generated by $\mu$ and $\Theta_\xi$ makes a wrong
prediction $x_t^{\Theta_\xi} \neq x_t$. Since
$\xi(x_t^{\Theta_\xi}|x_{<t}) \geq \xi(x_t|x_{<t})$, this implies
$\xi(x_t|x_{<t}) \leq \odt$. Hence $e_t^{\Theta_\xi} = 1 \leq
-\ln\xi(x_t|x_{<t})/\ln 2 = d_t/\ln 2$. If $\Theta_\xi$ makes a
correct prediction $e_t^{\Theta_\xi} = 0 \leq d_t/\ln 2$ is
obvious. Using (\ref{entropy}) this proves $E_\infty^{\Theta_\xi}  \leq
D_\infty/\ln 2  \leq  \log_2{w_\mu^{-1}}$.
A combinatoric argument given in Section \ref{secLower} shows that
there are $\M$ and $\mu\in\M$ with $E_\infty^{\Theta_\xi} \geq
\log_2|\M|$. This shows that the upper bound
$E_\infty^{\Theta_\xi} \leq \log_2|\M|$ for uniform $w$ is sharp.
From Theorem \ref{thErrBnd} we get the slightly weaker bound
$E_\infty^{\Theta_\xi}  \leq  2S_\infty  \leq  2D_\infty  \leq
2\ln{w_\mu^{-1}}$. For more complicated probabilistic
environments, where even the ideal informed system makes an
infinite number of errors, the theorem ensures that the error
regret $E_n^{\Theta_\xi} - E_n^{\Theta_\mu}$ is only of order
$\sqrt{E_n^{\Theta_\mu}}$.
\indxs{error}{regret}\indxs{error}{density}
The regret is quantified in terms of
the information content $D_n$ of $\mu$ (relative to $\xi$), or the
weight $w_\mu$ of $\mu$ in $\xi$. This ensures that the error
densities $E_n/n$ of both systems converge to each other.
Actually, the theorem ensures more, namely that the quotient
converges to 1, and also gives the speed of convergence
$E_n^{\Theta_\xi}/E_n^{\Theta_\mu}=1+O((E_n^{\Theta_\mu})^{-1/2})
\longrightarrow 1$ for $E_n^{\Theta_\mu}\to\infty$.
If we increase the first occurrence of $E_n^{\Theta_\mu}$ in the
theorem to $E_n^\Theta$ and the second to $E_n^{\Theta_\xi}$
we get the bound $E_n^\Theta\geq E_n^{\Theta_\xi}-
2\sqrt{E_n^{\Theta_\xi}S_n}$, which shows that {\em no} (causal)
predictor $\Theta$ whatsoever makes significantly less errors than
$\Theta_\xi$.
In Section \ref{secLower} we show that the third bound for
$E_n^{\Theta_\xi} - E_n^{\Theta_\mu}$ given in Theorem
\ref{thErrBnd} can in general not be improved, i.e.\ for every
predictor $\Theta$ (particularly $\Theta_\xi$) there exist $\M$
and $\mu\in\M$ such that the upper bound is essentially achieved. See
\tcite{Hutter:99errbnd} for some further discussion and bounds for
binary alphabet.

\subsection{Proof of Theorem \ref{thErrBnd}}\label{secPTErrBnd}
The first inequality in Theorem \ref{thErrBnd} has already been proven
(\ref{Emuopt}). For the second inequality, let us start more modestly
and try to find constants $A>0$ and $B>0$ that satisfy the linear
inequality
\beq\label{Eineq2}
  E_n^{\Theta_\xi} - E_n^{\Theta_\mu} \;\leq\;
  A Q_n + B S_n.
\eeq
If we could show
\beq\label{eineq2}
  e_t^{\Theta_\xi}(x_{<t}) - e_t^{\Theta_\mu}(x_{<t}) \;\leq\;
  A q_t(x_{<t}) + B s_t(x_{<t})
\eeq
for all $t \leq n$ and all $x_{<t}$, (\ref{Eineq2}) would follow
immediately by summation and the definition of $E_n$, $Q_n$ and $S_n$.
With the abbreviations
\beqn\label{xydef}
  \X=\{1,...,N\},\quad
  N=|\X|, \quad
  i=x_t, \quad
  y_i=\mu(x_t|x_{<t}), \quad
  z_i=\xi(x_t|x_{<t})
\eeqn
\beqn
  m=x_t^{\Theta_\mu}, \qquad
  s=x_t^{\Theta_\xi}
\eeqn
the various error functions can then be expressed by
$e_t^{\Theta_\xi}=1 - y_s$, $e_t^{\Theta_\mu}=1 - y_m$,
$q_t=1-\delta_{ms}$ and $s_t=\sum_i (y_i-z_i)^2$. Inserting this
into (\ref{eineq2}) we get
\beq\label{detineq}
  y_m\!-\!y_s \;\leq\;
  A[1\!-\!\delta_{ms}] + B\sum_{i=1}^N(y_i-z_i)^2.
\eeq
By definition of $x_t^{\Theta_\mu}$ and $x_t^{\Theta_\xi}$ we have
$y_m \geq y_i$ and $z_s \geq z_i$ for all $i$.
We prove a sequence of inequalities which show that
\beq\label{detineq2}
  B\sum_{i=1}^N(y_i-z_i)^2 + A[1\!-\!\delta_{ms}] - (y_m\!-\!y_s)
  \;\geq\; ...
\eeq
is positive for suitable $A \geq 0$ and $B \geq 0$, which proves
(\ref{detineq}). For $m=s$ (\ref{detineq2}) is obviously positive.
So we will assume $m \neq s$ in the following. From the square we
keep only contributions from $i=m$ and $i=s$.
\beqn
  ... \;\geq\;
  B[(y_m\!-\!z_m)^2+(y_s\!-\!z_s)^2] +
  A - (y_m\!-\!y_s)
  \;\geq\; ...
\eeqn
By definition of $y$, $z$, $\M$ and $s$ we have the constraints
$y_m + y_s \leq 1$, $z_m + z_s \leq 1$, $y_m \geq y_s \geq 0$
and $z_s \geq z_m \geq 0$. From the latter two it is easy to
see that the square terms (as a function of $z_m$ and $z_s$) are
minimized by $z_m=z_s=\odt(y_m+y_s)$. Together with the
abbreviation $x:=y_m-y_s$ we get
\beq\label{detineq4}
  ... \;\geq\;
  \odt B x^2 + A - x
  \;\geq\; ...
\eeq
(\ref{detineq4}) is quadratic in $x$ and minimized by
$x^* = {1\over B}$. Inserting $x^*$ gives
\beqn
  ... \;\geq\;
  A-{1\over 2B} \;\geq\; 0 \quad\mbox{for}\quad
   2AB\geq 1.
\eeqn
Inequality (\ref{Eineq2}) therefore holds for any $A > 0$,
provided we insert
$B = {1\over 2A}$. Thus we might
minimize the r.h.s.\ of (\ref{Eineq2}) w.r.t.\ $A$ leading to the upper bound
\beqn\label{detineq5}
E_n^{\Theta_\xi} - E_n^{\Theta_\mu} \;\leq\;
         \sqrt{2Q_nS_n}
\qquad\mbox{for}\qquad A^2={S_n\over 2Q_n}
\eeqn
which is the first bound in Theorem \ref{thErrBnd}.
For the second bound we have to show $Q_n\leq E_n^{\Theta_\xi} +
E_n^{\Theta_\mu}$, which follows by summation from $q_t\leq
e_t^{\Theta_\xi} + e_t^{\Theta_\mu}$, which is equivalent to
$1-\delta_{ms}\leq 1-y_s+1-y_m$, which holds for $m=s$ as well as
$m\neq s$.
For the third bound we have to prove
\beq\label{detineq6}
  \sqrt{2(E_n^{\Theta_\xi}\!+\! E_n^{\Theta_\mu})S_n} - S_n \;\leq\;
  \sqrt{4E_n^{\Theta_\mu}S_n+S_n^2}.
\eeq
If we square both sides of this expressions and simplify we just
get the second bound. Hence, the second bound implies
(\ref{detineq6}).
The last inequality in Theorem \ref{thErrBnd} is a simple triangle
inequality. This completes the proof of Theorem \ref{thErrBnd}.
\qed

Note that also the third bound implies the second one:
\beqn
  E_n^{\Theta_\xi}-E_n^{\Theta_\mu} \;\leq\;
  \sqrt{2(E_n^{\Theta_\xi}\!+\! E_n^{\Theta_\mu})S_n}
  \quad\Leftrightarrow\quad
  (E_n^{\Theta_\xi}\!-\!E_n^{\Theta_\mu})^2 \;\leq\;
  2(E_n^{\Theta_\xi}\!+\! E_n^{\Theta_\mu})S_n
  \quad\Leftrightarrow
\eeqn\vspace{-2ex}
\beqn
  \Leftrightarrow\quad
  (E_n^{\Theta_\xi}\!-\!E_n^{\Theta_\mu}\!-\!S_n)^2 \;\leq\;
  4E_n^{\Theta_\mu}S_n+S_n^2
  \quad\Leftrightarrow\quad
  E_n^{\Theta_\xi}-E_n^{\Theta_\mu}-S_n \;\leq\;
  \sqrt{4E_n^{\Theta_\mu}S_n+S_n^2}
\eeqn
where we only have used $E_n^{\Theta_\xi}\geq E_n^{\Theta_\mu}$.
Nevertheless the bounds are not equal.

\subsection{General Loss Function}\label{secLoss}
A prediction is very often the basis for some decision. The
decision results in an action, which itself leads to some reward
or loss. If the action itself can influence the environment we
enter the domain of acting agents which has been analyzed in the
context of universal probability \citein{Hutter:01aixi}. To stay
in the framework of (passive) prediction we have to assume that
the action itself does not influence the environment. Let
$\ell_{x_t y_t} \in \Set{R}$ be the received loss when taking
action $y_t \in \Y$ and $x_t\in\X$ is the $t^{th}$ symbol of the
sequence. We make the assumption that $\ell$ is bounded.
Without loss of generality we normalize $\ell$ by linear scaling
such that $0 \leq \ell_{x_t y_t} \leq 1$. For instance, if we make
a sequence of \idx{weather forecasts} $\X = \{$sunny, rainy$\}$
and base our decision, whether to take an umbrella or wear
sunglasses $\Y = \{$umbrella, sunglasses$\}$ on it, the action of
taking the umbrella or wearing sunglasses does not influence the
future weather (ignoring the \idx{butterfly effect}). The losses
might be

\begin{center}
\begin{tabular}{|c|c|c|}\hline
  Loss & sunny & rainy        \\\hline
  umbrella & 0.1 & 0.3   \\\hline
  sunglasses & 0.0 & 1.0         \\\hline
\end{tabular}
\end{center}

\noindent Note the loss assignment even when making the
right decision to take an umbrella when it rains because sun is
still preferable to rain.

In many cases the prediction of $x_t$ can be identified or is already
the action $y_t$. The forecast {\em sunny} can be identified
with the action {\em wear sunglasses}, and {\em rainy} with {\em
take umbrella}. $\X \equiv\Y$ in these cases. The error
assignment of the previous subsections falls into this class
together with a special loss function. It assigns unit loss to an
erroneous prediction ($\ell_{x_t y_t} = 1$ for $x_t \neq y_t$) and
no loss to a correct prediction ($\ell_{x_t x_t} = 0$).

\indxs{Bayes-optimal}{prediction} For convenience we name an
action a prediction in the following, even if $\X \neq\Y$. The
true probability of the next symbol being $x_t$, given $x_{<t}$,
is $\mu(x_t|x_{<t})$. The expected loss when predicting $y_t$ is
$\E_t[\ell_{x_t y_t}]$.
\indxs{minimize}{loss} The goal is to minimize the expected loss.
More generally we define the $\Lambda_\rho$ prediction scheme
\beq\label{xlrdef}
  y_t^{\Lambda_\rho} \;:=\;
  \arg\min_{y_t\in\Y}\sum_{x_t}\rho(x_t|x_{<t})\ell_{x_t y_t}
\eeq
which minimizes the $\rho$-expected loss.$\!$ %
\footnote{$\arg\min_y(\cdot)$ is defined as the $y$ which
minimizes the argument. A tie is broken arbitrarily. In general,
the prediction space $\Y$ is allowed to differ from $\X$. If $\Y$
is finite, then $y_t^{\Lambda_\rho}$ always exists. For an infinite
action space $\Y$ we assume that a minimizing $y_t^{\Lambda_\rho}
\in \Y$ exists, although even this assumption may be removed.} As
the true distribution is $\mu$, the actual $\mu$-expected loss
when $\Lambda_\rho$ predicts the $t^{th}$ symbol and the total
$\mu$-expected loss in the first $n$ predictions are
\beq\label{rholoss}
  l_t^{\Lambda_\rho}(x_{<t}) \;:=\;
  \E_t \ell_{x_t y_t^{\Lambda_\rho}}
  \quad,\quad
  L_n^{\Lambda_\rho} \;:=\; \sum_{t=1}^n
  \E_{<t} l_t^{\Lambda_\rho}(x_{<t}).
\eeq
Let $\Lambda$ be {\em any} (causal) prediction scheme
(deterministic or probabilistic does not matter) with no
constraint at all, predicting {\em any} $y_t^\Lambda \in \Y$ with
losses $l_t^{\Lambda}$ and $L_n^{\Lambda}$ similarly defined as
(\ref{rholoss}). If $\mu$ is known, $\Lambda_\mu$ is obviously the
best prediction scheme in the sense of achieving minimal expected
loss
\beq\label{Lmuopt}
  L_n^{\Lambda_\mu} \;\leq\;L_n^{\Lambda} \qmbox{for any}
  \Lambda.
\eeq
The following loss bound for the universal
$\Lambda_\xi$ predictor is proven
\citein{Hutter:02spupper}.
\beq\label{th3}\label{thGLoss}\label{thULoss}
  0 \;\leq\; L_n^{\Lambda_\xi}-L_n^{\Lambda_\mu} \;\leq\;
  D_n+\sqrt{4L_n^{\Lambda_\mu}D_n+D_n^2} \;\leq\;
  2D_n+2\sqrt{L_n^{\Lambda_\mu}D_n}.
\eeq
The loss bounds have the same form as the error bounds when
substituting $S_n\leq D_n$ in Theorem \ref{thErrBnd}. For a
comparison to Merhav's and Feder's \citey{Merhav:98} loss bound,
see \tcite{Hutter:02spupper}. Replacing $D_n$ by $S_n$ or $V_n$ in (\ref{th3}) gives an invalid bound, so the general
bound is slightly weaker. For instance, for $\X=\{0,1\}$,
$\ell_{00}=\ell_{11}=0$, $\ell_{10}=1$, $\ell_{01}=c<{1\over 4}$,
$\mu(1)=0$, $\nu(1)=2c$, and $w_\mu=w_\nu=\odt$ we get $\xi(1)=c$,
$s_1=2c^2$, $y_1^{\Lambda_\mu}=0$,
$l_1^{\Lambda_\mu}=\ell_{00}=0$, $y_1^{\Lambda_\xi}=1$,
$l_1^{\Lambda_\xi}=\ell_{01}=c$, hence
$L_1^{\Lambda_\xi}-L_1^{\Lambda_\mu}=c \;\not\leq\;
4c^2=2S_1+2\sqrt{L_1^{\Lambda_\mu}S_1}$. Example loss functions
including the absolute, square, logarithmic, and Hellinger loss
are discussed in \tcite{Hutter:02spupper}. Instantaneous error/loss
bounds can also be proven:
\beqn
 e_t^{\Theta_\xi}(x_{<t})-e_t^{\Theta_\mu}(x_{<t}) \;\leq\; \sqrt{2s_t(x_{<t})}
 ,\quad
 l_t^{\Lambda_\xi}(x_{<t})-l_t^{\Lambda_\mu}(x_{<t}) \;\leq\; \sqrt{2d_t(x_{<t})}
 .
\eeqn

\section{Application to Games of Chance}\label{secGames}
\index{application!games of chance}\index{games!of chance}

This section applies
the loss bounds to games of chance, defined as a sequence of bets,
observations, and rewards.
After a brief introduction in Section \ref{secGCIntro}
we show in Section \ref{secGC} that if there is a profitable
scheme at all, asymptotically the universal $\Lambda_\xi$ scheme
will also become profitable. We bound the time needed to reach the
winning zone. It is proportional to the relative entropy of $\mu$
and $\xi$ with a factor depending on the profit range and the
average profit.
Section \ref{secGCEx} presents a numerical example and
Section \ref{secGCITI} attempts to give an information theoretic
interpretation of the result.

\subsection{Introduction}\label{secGCIntro}
\index{stock market} \index{portfolio} \index{investment}
\index{profit} \index{trader} \index{money} Consider investing in
the stock market. At time $t$ an amount of money $s_t$ is invested
in portfolio $y_t$, where we have access to past knowledge
$x_{<t}$ (e.g.\ charts). After our choice of investment we receive
new information $x_t$, and the new portfolio value is $r_t$. The
best we can expect is to have a probabilistic model $\mu$ of the
behavior of the stock-market. The goal is to maximize the net
$\mu$-expected profit $p_t = r_t - s_t$. Nobody knows $\mu$, but
the assumption of all traders is that there {\em is} a computable,
profitable $\mu$ they try to find or approximate. From Theorem
\ref{thConv} we know that Levin's universal prior
$\xi_U(x_t|x_{<t})$ converges to any computable $\mu(x_t|x_{<t})$
with probability 1. If there is a computable, asymptotically
profitable trading scheme at all, the $\Lambda_\xi$ scheme should
also be profitable in the long run. To get a practically useful,
computable scheme we have to restrict $\M$ to a finite set of
computable distributions, e.g.\ with bounded Levin complexity $Kt$
\cite{Li:97}. Although convergence of $\xi$ to $\mu$ is pleasing,
what we are really interested in is whether $\Lambda_\xi$ is
asymptotically profitable and how long it takes to become
profitable. This will be explored in the following.

\subsection{Games of Chance}\label{secGC}
\index{bet} \index{time!to win}\indxs{winning}{strategy} We use
the loss bound (\ref{thGLoss}) to estimate the time needed to
reach the winning threshold when using $\Lambda_\xi$ in a game of
chance. We assume a game (or a sequence of possibly correlated
games) which allows a sequence of bets and observations. In step
$t$ we bet, depending on the history $x_{<t}$, a certain amount of
money $s_t$, take some action $y_t$, observe outcome $x_t$, and
receive reward $r_t$. Our profit, which we want to maximize, is
$p_t=r_t-s_t \in [p_{min},p_{max}]$, where $[p_{min},p_{max}]$ is
the [minimal,maximal] profit per round and
$p_\Delta:=p_{max}-p_{min}$ the profit range. The
loss, which we want to minimize, can be defined as the negative
scaled profit, $\ell_{x_t y_t} = (p_{max}-p_t)/p_\Delta \in
[0,1]$. The probability of outcome $x_t$, possibly depending on
the history $x_{<t}$, is $\mu(x_t|x_{<t})$. The total
$\mu$-expected profit when using scheme $\Lambda_\rho$ is
$P_n^{\Lambda_\rho} = np_{max} - p_\Delta L_n^{\Lambda_\rho}$. If
we knew $\mu$, the optimal strategy to maximize our expected
profit is just $\Lambda_\mu$. We assume $P_n^{\Lambda_\mu} > 0$
(otherwise there is no winning strategy at all, since
$P_n^{\Lambda_\mu} \geq P_n^\Lambda\,\forall\Lambda$). Often we
are not in the favorable position of knowing $\mu$, but we know
(or assume) that $\mu\in\M$ for some $\M$, for instance that $\mu$
is a computable probability distribution. From bound
(\ref{thGLoss}) we see that the\indxs{profit}{average} average
profit per round $\bar p_n^{\Lambda_\xi} := {1\over
n}P_n^{\Lambda_\xi}$ of the universal $\Lambda_\xi$ scheme
converges to the average profit per round $\bar p_n^{\Lambda_\mu}
:= {1\over n}P_n^{\Lambda_\mu}$ of the optimal informed scheme,
i.e.\ asymptotically we can make the same money even without
knowing $\mu$, by just using the universal $\Lambda_\xi$ scheme.
Bound (\ref{thGLoss}) allows us to lower bound the universal
profit $P_n^{\Lambda_\xi}$
\beq\label{pbnd}
  P_n^{\Lambda_\xi} \;\geq\; P_n^{\Lambda_\mu} -
  p_\Delta D_n-\sqrt{4(n p_{max}\!-\!P_n^{\Lambda_\mu})
  p_\Delta D_n+p_\Delta^2D_n^2}.
\eeq
\indxs{maximal}{profit}The time needed for $\Lambda_\xi$ to
perform well can also be estimated. An interesting quantity is the
expected number of rounds needed to reach the winning zone. Using
$P_n^{\Lambda_\mu} > 0$ one can show that the r.h.s.\ of
(\ref{pbnd}) is positive if, and only if
\beq\label{pwin}
  n \;>\; {2p_\Delta(2p_{max}\!-\!\bar p_n^{\Lambda_\mu}) \over
  (\bar p_n^{\Lambda_\mu})^2} \!\cdot\! D_n.
\eeq

\ftheorem{thWin}{Time to Win}{
\index{time!to win}%
Let there be sequences $x_1x_2...$ over a finite alphabet
$\X$ drawn with probability $\mu(x_{1:n})$ for the first
$n$ symbols. In step $t$ we make a bet, depending on the history
$x_{<t}$, take some action $y_t$, and observe outcome $x_t$. Our
net profit is $p_t \in [p_{max} - p_\Delta,p_{max}]$. The
$\Lambda_\rho$-system (\ref{xlrdef}) acts as to maximize the
$\rho$-expected profit. $P_n^{\Lambda_\rho}$ is the total and
$\bar p_n^{\Lambda_\rho} = {1\over n}P_n^{\Lambda_\rho}$ is the average
expected profit of the first $n$ rounds. For the universal
$\Lambda_\xi$ and for the optimal informed $\Lambda_\mu$
prediction scheme the following holds:
\beqn
\begin{array}{rl}
   i) & \bar p_n^{\Lambda_\xi} \;=\; \bar p_n^{\Lambda_\mu}-O(n^{-1/2})
        \longrightarrow \bar p_n^{\Lambda_\mu}
        \quad\mbox{for}\quad n\to\infty \\[1ex]
  ii) & n \;>  \Big({2p_\Delta\over
        \bar p_n^{\Lambda_\mu}}\Big)^{\!2}
        \!\cdot\!b_\mu \quad\wedge\quad \bar p_n^{\Lambda_\mu}>0
        \quad\Longrightarrow\quad \bar p_n^{\Lambda_\xi}>0
\end{array}
\eeqn
where $b_\mu=\ln w_\mu^{-1}$ with $w_\mu$ being the weight
(\ref{xidefsp}) of $\mu$ in $\xi$ in the discrete case (and $b_\mu$
as in Theorem \ref{thCEB} in the continuous case).
}

\noindent By dividing (\ref{pbnd}) by $n$ and using $D_n \leq b_\mu$
(\ref{entropy}) we see that the leading order of
$\bar p_n^{\Lambda_\xi} - \bar p_n^{\Lambda_\mu}$ is bounded by
$\sqrt{4p_\Delta p_{max}b_\mu/n}$, which proves $(i)$. The
condition in $(ii)$ is actually a weakening of (\ref{pwin}).
$P_n^{\Lambda_\xi}$ is trivially positive for $p_{min} > 0$, since
in this wonderful case {\em all} profits are positive. For
negative $p_{min}$ the condition of $(ii)$ implies
(\ref{pwin}), since $p_\Delta > p_{max}$, and (\ref{pwin}) implies
positive (\ref{pbnd}), i.e.\ $P_n^{\Lambda_\xi} > 0$, which proves
$(ii)$.

If a winning strategy $\Lambda$ with $\bar p_n^\Lambda > \eps >0$
exists, then $\Lambda_\xi$ is asymptotically also a winning
strategy with the same average profit.

\subsection{Example}\label{secGCEx}
\indxs{dice}{example}
Let us consider a game with two dice, one with two black and
four white faces, the other with four black and two white faces.
The dealer who repeatedly throws the dice uses one or the other
die according to some deterministic rule, which correlates the
throws (e.g.\ the first die could be used in round $t$ iff the $t^{th}$
digit of $\pi$ is 7). We can bet on black or white;
the stake $s$ is 3\$ in every round; our return $r$ is 5\$ for
every correct prediction.

The profit is $p_t = r\delta_{x_ty_t} - s$. The coloring of the
dice and the selection strategy of the dealer unambiguously
determine $\mu$. $\mu(x_t|x_{<t})$ is ${1\over 3}$ or ${2\over 3}$
depending on which die has been chosen. One should bet on the more
probable outcome. If we knew $\mu$ the expected profit per round
would be $\bar p_n^{\Lambda_\mu} = p_n^{\Lambda_\mu} = {2\over 3}r
- s = {1\over 3}\$>0$. If we don't know $\mu$ we should use
Levin's universal prior with $D_n \leq b_\mu = K(\mu) \cdot \ln
2$, where $K(\mu)$ is the length of the shortest program coding
$\mu$ (see Section \ref{secPCM}). Then we know  that betting on
the outcome with higher $\xi$ probability leads asymptotically to
the same profit (Theorem \ref{thWin}$(i)$) and $\Lambda_\xi$
reaches the \indxs{winning}{threshold} winning threshold no later
than $n_{thresh} = 900\ln 2 \cdot K(\mu)$ (Theorem
\ref{thWin}$(ii)$) or sharper $n_{thresh} = 330\ln 2 \cdot K(\mu)$
from (\ref{pwin}), where $p_{max} = r-s = 2\$ $ and $p_\Delta = r
= 5\$ $ have been used.

\index{strategy!die selection}
If the die selection strategy reflected in $\mu$ is not too
complicated, the $\Lambda_\xi$ prediction system reaches the
winning zone after a few thousand rounds. The number of rounds is
not really small because the expected profit per round is one
order of magnitude smaller than the return. This leads to a
constant of two orders of magnitude size in front of $K(\mu)$.
Stated otherwise, it is due to the large stochastic noise, which
makes it difficult to extract the signal, i.e.\ the structure of
the rule $\mu$ (see next subsection). Furthermore, this is only a
bound for the turnaround value of $n_{thresh}$. The true expected
turnaround $n$ might be smaller. However, for every game for which
there exists a computable winning strategy with $\bar
p_n^\Lambda >\eps > 0$, $\Lambda_\xi$ is guaranteed to get
into the winning zone for some $n \sim K(\mu)$.

\subsection{Information-Theoretic Interpretation}\label{secGCITI}
We try to give an intuitive explanation of Theorem
\ref{thWin}$(ii)$. We know that $\xi(x_t|x_{<t})$ converges to
$\mu(x_t|x_{<t})$ for $t\to\infty$. In a sense $\Lambda_\xi$
learns $\mu$ from past data $x_{<t}$. The \idx{information}
content in $\mu$ relative to $\xi$ is $D_\infty/\ln 2 \leq
b_\mu/\ln 2$. One might think of a Shannon-Fano prefix code
of $\nu \in \M$ of length $^\lceil b_\nu/\ln 2^\rceil$,
which exists since the Kraft inequality $\sum_\nu 2^{-^\lceil
b_\nu/\ln 2^\rceil} \leq \sum_\nu w_\nu\leq 1$ is
satisfied. $b_\mu/\ln 2$ bits have to be learned before
$\Lambda_\xi$ can be as good as $\Lambda_\mu$. In the worst case,
the only information conveyed by $x_t$ is in form of the received
profit $p_t$. Remember that we always know the profit $p_t$ before
the next cycle starts.

\indxs{signal}{amplitude}\indxs{information}{transmission}
Assume that the distribution of the profits in the interval
$[p_{min},p_{max}]$ is mainly due to noise, and there is only a
small informative signal of amplitude $\bar p_n^{\Lambda_\mu}$. To
reliably determine the sign of a signal of amplitude $\bar
p_n^{\Lambda_\mu}$, disturbed by noise of amplitude $p_\Delta$, we
have to resubmit a bit $O((p_\Delta/\bar p_n^{\Lambda_\mu})^2)$
times (this reduces the standard deviation below the signal
amplitude $\bar p_n^{\Lambda_\mu}$). To learn $\mu$, $b_\mu/\ln 2$
bits have to be transmitted, which requires $n \geq
O((p_\Delta/\bar p_n^{\Lambda_\mu})^2) \cdot b_\mu/\ln 2$ cycles.
This expression coincides with the condition in $(ii)$.
Identifying the signal amplitude with $\bar p_n^{\Lambda_\mu}$ is
the weakest part of this consideration, as we have no argument why
this should be true. It may be interesting to make the analogy
more rigorous, which may also lead to a simpler proof of $(ii)$
not based on bounds (\ref{thULoss}) with their rather
complex proofs.

\section{Optimality Properties}\label{secLower}
\index{optimality properties}

In this section we discuss the quality of the universal predictor
and the bounds.
In Section \ref{secLEB} we show that there are $\M$ and $\mu\in\M$
and weights $w_\nu$ such that the derived error bounds are tight.
This shows that the error bounds cannot be improved in general.
In Section \ref{secPareto} we show Pareto-optimality of
$\xi$ in the sense that there is no other predictor which performs
at least as well in all environments $\nu\in\M$ and strictly
better in at least one. Optimal predictors can always be based on
mixture distributions $\xi$.
This still leaves open how to choose the weights. In Section
\ref{subsecWeights} we give an Occam's razor argument that the
choice $w_\nu=2^{-K(\nu)}$, where $K(\nu)$ is the length of the
shortest program describing $\nu$ is optimal.

\subsection{Lower Error Bound}\label{secLEB}
\index{bound!lower}\index{error bound!lower}
We want to show that there exists a class $\M$ of distributions
such that {\em any} predictor $\Theta$ ignorant of the
distribution $\mu\in\M$ from which the observed sequence is
sampled must make some minimal additional number of errors as
compared to the best informed predictor $\Theta_\mu$.

For deterministic environments a lower bound can easily be
obtained by a combinatoric argument. Consider a class $\M$
containing $2^n$ binary sequences such that each prefix of length
$n$ occurs exactly once. Assume any deterministic predictor
$\Theta$ (not knowing the sequence in advance), then for every
prediction $x_t^\Theta$ of $\Theta$ at times $t \leq n$ there
exists a sequence with opposite symbol $x_t = 1 - x_t^\Theta$.
Hence, $E_\infty^{\Theta} \geq E_n^{\Theta} = n = \log_2|\M|$ is a
lower worst case bound for every predictor $\Theta$, (this
includes $\Theta_\xi$, of course). This shows that the upper bound
$E_\infty^{\Theta_\xi} \leq \log_2|\M|$ for uniform $w$ obtained
in the discussion after Theorem \ref{thErrBnd} is sharp. In the
general probabilistic case we can show by a similar argument that
the upper bound of Theorem \ref{thErrBnd} is sharp for
$\Theta_\xi$ and ``static'' predictors, and sharp within a factor
of 2 for general predictors. We do not know whether the factor two
gap can be closed.
\index{bound!tight}\index{error bound!tight}
\index{bound!sharp}\index{error bound!sharp}

\ftheorem{thLow}{Lower Error Bound}{
For every $n$ there is an $\M$ and $\mu\in\M$ and
weights $w_\nu$ such that
\beqn
  (i) \qquad\qquad e_t^{\Theta_\xi}-e_t^{\Theta_\mu} \;=\; \sqrt{2s_t}
  \qmbox{and}
  E_n^{\Theta_\xi} -E_n^{\Theta_\mu} \;=\;
  S_n + \sqrt{4E_n^{\Theta_\mu}S_n+S_n^2}
\eeqn
where $E_n^{\Theta_\xi} $and $E_n^{\Theta_\mu}$ are the total
expected number of errors of $\Theta_\xi$ and $\Theta_\mu$, and
$s_t$ and $S_n$ are defined in (\ref{stmuxi}). More generally, the
equalities hold for {\em any ``static''} deterministic predictor
$\theta$ for which $y_t^\Theta$ is independent of $x_{<t}$.
For every $n$ and {\em arbitrary} deterministic predictor
$\Theta$, there exists an $\M$ and $\mu\in\M$
such that
\beqn
  (ii) \qquad e_t^\Theta-e_t^{\Theta_\mu} \;\geq\; \odt\sqrt{2s_t(x_{<t})}
  \qmbox{and}
  E_n^\Theta -E_n^{\Theta_\mu} \;\geq\;
  \odt[S_n + \sqrt{4E_n^{\Theta_\mu}S_n+S_n^2}]
\eeqn
}

\paradot{Proof} $(i)$ The proof parallels and generalizes the deterministic
case. Consider a class $\M$ of $2^n$ distributions (over binary
alphabet) indexed by $a \equiv a_1...a_n \in\{0,1\}^n$. For each
$t$ we want a distribution with posterior probability
$\odt(1+\eps)$ for $x_t=1$ and one with posterior probability
$\odt(1-\eps)$ for $x_t=1$ independent of the past $x_{<t}$ with
$0<\eps\leq\odt$. That is
\beqn
  \mu_a(x_1...x_n) = \mu_{a_1}(x_1)\cdot...\cdot\mu_{a_n}(x_n),
  \qmbox{where}
  \mu_{a_t}(x_t) = \left\{ {\odt(1+\eps) \quad\mbox{for}\quad x_t=a_t \atop
                           \odt(1-\eps) \quad\mbox{for}\quad x_t\neq a_t} \right.
\eeqn
We are not interested in predictions beyond time $n$ but for
completeness we may define $\mu_a$ to assign probability 1 to
$x_t=1$ for all $t>n$. If $\mu =\mu_a$, the informed scheme
$\Theta_\mu$ always predicts the bit which has highest $\mu$-probability, i.e.\ $y_t^{\Theta_\mu} =a_t$
\beqn
  \Longrightarrow\quad e_t^{\Theta_\mu} = 1-\mu_{a_t}(y_t^{\Theta_\mu})=\odt(1-\eps)
  \quad\Longrightarrow\quad E_n^{\Theta_\mu} = \textstyle{n\over
  2}(1-\eps).
\eeqn
Since $E_n^{\Theta_\mu}$ is the same for all $a$ we seek to
maximize $E_n^\Theta $ for a given predictor $\Theta$ in the
following. Assume $\Theta$ predicts $y_t^\Theta$ (independent of
history $x_{<t}$). Since we want lower bounds we seek a worst
case $\mu$. A success $y_t^\Theta = x_t$ has lowest possible
probability $\odt(1-\eps)$ if $a_t=1-y_t^\Theta$.
\beqn
  \Longrightarrow\quad e_t^\Theta  = 1-\mu_{a_t}(y_t^{\Theta})=\odt(1+\eps)
  \quad\Longrightarrow\quad E_n^\Theta  = \textstyle{n\over
  2}(1+\eps).
\eeqn
So we have $e_t^\Theta  - e_t^{\Theta_\mu}  =  \eps$ and
$E_n^\Theta  - E_n^{\Theta_\mu}  =  n\eps$ for the regrets. We
need to eliminate $n$ and $\eps$ in favor of $s_t$, $S_n$, and
$E_n^{\Theta_\mu}$. If we assume uniform weights
$w_{\mu_a} = 2^{-n}$ for all $\mu_a$ we get
\beqn
  \xi(x_{1:n}) \;=\; \sum_a w_{\mu_a} \mu_a(x_{1:n}) \;=\;
  2^{-n}\prod_{t=1}^n\sum_{a_t\in\{0,1\}}\mu_{a_t}(x_t) \;=\;
  2^{-n}\prod_{t=1}^n 1 \;=\; 2^{-n},
\eeqn
i.e.\ $\xi$ is an unbiased Bernoulli sequence
\indxs{Bernoulli}{sequence}
($\xi(x_t|x_{<t})=\odt$).
\beqn
  \Longrightarrow\quad s_t(x_{<t}) \;=\;
  \sum_{x_t}(\odt-\mu_{a_t}(x_t))^2 \;=\; \odt\eps^2
  \qmbox{and}
  S_n=\textstyle{n\over 2}\eps^2.
\eeqn
So we have $\eps = \sqrt{2s_t}$ which proves the instantaneous
regret formula $e_t^\Theta  - e_t^{\Theta_\mu} = \sqrt{2s_t}$ for
static $\Theta$. Inserting $\eps = \sqrt{{2\over n}S_n}$ into
$E_n^{\Theta_\mu}$ and solving w.r.t.\ $\sqrt{2n}$ we get
$\sqrt{2n} = \sqrt{S_n}+\sqrt{4E_n^{\Theta_\mu}+S_n}$. So we
finally get
\beqn
  E_n^\Theta -E_n^{\Theta_\mu} \;=\; n\eps \;=\; \sqrt{S_n}\sqrt{2n} \;=\;
  S_n + \sqrt{4E_n^{\Theta_\mu}S_n+S_n^2}
\eeqn
which proves the total regret formula in $(i)$ for static
$\Theta$. We can choose\footnote{This choice may be made unique by
slightly non-uniform $w_{\mu_a} =
\prod_{t=1}^n[\odt+(\odt-a_t)\delta]$ with $\delta\ll 1$.}
$y_t^{\Theta_\xi}\equiv 0$ to be a static predictor. Together this
shows $(i)$.

$(ii)$ For non-static predictors, $a_t=1-y_t^\Theta$ in the proof
of $(i)$ depends on $x_{<t}$, which is not allowed. For general,
but fixed $a_t$ we have
$e_t^\Theta(x_{<t})=1-\mu_{a_t}(y_t^\Theta)$. This quantity may
assume any value between $\odt(1-\eps)$ and $\odt(1+\eps)$, when
averaged over $x_{<t}$, and is, hence of little direct help. But
if we additionally average the result also over all environments
$\mu_a$, we get
\beqn
  < E_n^\Theta >_a\;
  \;=\; < \sum_{t=1}^n \E[e_t^\Theta(x_{<t})] >_a\;
  \;=\; \sum_{t=1}^n \E[< e_t^\Theta(x_{<t}) >_a]
  \;=\; \sum_{t=1}^n \E[\odt] = \odt n
\eeqn
whatever $\Theta$ is chosen: a sort of No-Free-Lunch theorem
\cite{Wolpert:97}, stating that on {\em uniform} average all
predictors perform equally well/bad. The expectation of
$E_n^\Theta$ w.r.t.\ $a$ can only be $\odt n$ if
$E_n^\Theta\geq\odt n$ for some $a$. Fixing such an $a$ and
choosing $\mu=\mu_a$ we get $E_n^\Theta -E_n^{\Theta_\mu} \geq
\odt n\eps = \odt[S_n + \sqrt{4E_n^{\Theta_\mu}S_n+S_n^2}]$,
and similarly $e_n^\Theta-e_n^{\Theta_\mu} \geq \odt\eps =
\odt\sqrt{2s_t(x_{<t})}$. \qed

\indxs{asymptotic}{optimality} Since for binary alphabet $s_t=\odt
a^2_t$, Theorem \ref{thLow} also holds with $s_t$ replaced by $\odt a^2_t$
and $S_n$ replaced by $V_n$.
Since $d_t/s_t=1+O(\eps^2)$ we have $D_n/S_n \to 1$ for $\eps \to
0$. Hence the error bound of Theorem \ref{thErrBnd} with $S_n$
replaced by $D_n$ is asymptotically tight for
$E_n^{\Theta_\mu}/D_n\to\infty$ (which implies $\eps \to 0$). This
shows that without restrictions on the loss function which exclude
the error loss, the loss bound (\ref{thULoss}) can also not be
improved. Note that the bounds are tight even when $\M$ is
restricted to Markov or i.i.d.\ environments, since the presented
counterexample is i.i.d.

A set $\M$ independent of $n$ leading to a good (but not tight)
lower bound is $\M = \{\mu_1,\mu_2\}$ with $\mu_{1/2}(1|x_{<t}) =
\odt \pm \eps_t$ with $\eps_t = \min\{\odt,\sqrt{\ln
w_{\mu_1}^{-1}}/\sqrt t\ln t\}$.
For $w_{\mu_1} \ll w_{\mu_2}$ and $n\to\infty$ one can show that
$E_n^{\Theta_\xi} - E_n^{\Theta_{\mu_1}} \sim {1\over\ln
n}\sqrt{E_n^{\Theta_\mu}\ln w_{\mu_1}^{-1}}$.

Unfortunately there are many important special cases for which the
loss bound (\ref{th3}) is not tight. For continuous $\Y$ and
logarithmic or quadratic loss function, for instance, one can show
that the regret $L_\infty^{\Lambda_\xi}-L_\infty^{\Lambda_\mu}\leq
\ln w_\mu^{-1}<\infty$ is finite \cite{Hutter:02spupper}. For
arbitrary loss function, but $\mu$ bounded away from certain
critical values, the regret is also finite. For instance, consider
the special error-loss, binary alphabet, and
$|\mu(x_t|x_{<t})-\odt|>\eps$ for all $t$ and $x$. $\Theta_\mu$
predicts $0$ if $\mu(0|x_{<t})>\odt$. If also
$\xi(0|x_{<t})>\odt$, then $\Theta_\xi$ makes the same prediction
as $\Theta_\mu$, for $\xi(0|x_{<t})<\odt$ the predictions differ.
In the latter case $|\xi(0|x_{<t})-\mu(0|x_{<t})|>\eps$.
Conversely for $\mu(0|x_{<t})<\odt$. So in any case
$e_t^{\Theta_\xi}-e_t^{\Theta_\mu}\leq
{1\over\eps^2}[\xi(x_t|x_{<t})-\mu(x_t|x_{<t})]^2$. Using
(\ref{rhoerr}) and Theorem \ref{thConv} we see that
$E_\infty^{\Theta_\xi}-E_\infty^{\Theta_\mu}\leq {1\over\eps^2}\ln
w_\mu^{-1}<\infty$ is finite too. Nevertheless, Theorem
\ref{thLow} is important as it tells us that bound (\ref{thULoss})
can only be strengthened by making further assumptions on $\ell$
or $\M$.

\subsection{Pareto Optimality of $\xi$}\label{secPareto}
\index{Pareto-optimality}\index{optimality!Pareto}
In this subsection we want to establish a different kind of
optimality property of $\xi$. Let $\F(\mu,\rho)$ be any of the
performance measures of $\rho$ relative to $\mu$ considered in the
previous sections (e.g.\ $s_t$, or $D_n$, or $L_n$, ...). It is
easy to find $\rho$ more tailored towards $\mu$ such that
$\F(\mu,\rho)<\F(\mu,\xi)$. This improvement may be achieved by
increasing $w_\mu$, but probably at the expense of increasing $\F$
for other $\nu$, i.e.\ $\F(\nu,\rho)>\F(\nu,\xi)$ for some
$\nu\in\M$. Since we do not know $\mu$ in advance we may ask whether
there exists a $\rho$ with better or equal performance for {\em
all} $\nu \in\M$ and a strictly better performance for one
$\nu \in\M$. This would clearly render $\xi$ suboptimal w.r.t.\
to $\F$. We show that there is no such $\rho$ for most performance
measures studied in this work.

\fdefinition{defPareto}{Pareto Optimality}{
Let $\F(\mu,\rho)$ be any performance measure of $\rho$ relative to
$\mu$. The universal prior $\xi$ is called Pareto-optimal w.r.t.\
$\F$ if there is no $\rho$ with $\F(\nu,\rho)\leq\F(\nu,\xi)$
for all $\nu \in \M$ and strict inequality for at least one
$\nu$.
}

\ftheorem{thPareto}{Pareto Optimality}{
The universal prior $\xi$ is Pareto-optimal w.r.t.\ the
instantaneous and total squared distances $s_t$ and $S_n$
(\ref{stmuxi}), entropy distances $d_t$ and $D_n$ (\ref{dtmuxi}),
errors $e_t$ and $E_n$ (\ref{rhoerr}), and losses $l_t$ and $L_n$
(\ref{rholoss}).
}

\paradot{Proof} We first prove Theorem \ref{thPareto} for the
instantaneous expected loss $l_t$. We need the more general
$\rho$-expected instantaneous losses
\beq\label{mxexloss}
  l_{t\rho}^{\Lambda}(x_{<t})\;:=\;\sum_{x_t}\rho(x_t|x_{<t})\ell_{x_ty_t^\Lambda}
\eeq
for a predictor $\Lambda$.
We want to arrive at a contradiction by assuming that $\xi$ is not
Pareto-optimal, i.e.\ by assuming the existence of a
predictor\footnote{According to Definition \ref{defPareto} we
should look for a $\rho$, but for each deterministic predictor
$\Lambda$ there exists a $\rho$ with $\Lambda=\Lambda_\rho$.}
$\Lambda$ with $l_{t\nu}^\Lambda\leq l_{t\nu}^{\Lambda_\xi}$ for all
$\nu\in\M$ and strict inequality for some $\nu$. Implicit to this
assumption is the assumption that $l_{t\nu}^\Lambda $ and
$l_{t\nu}^{\Lambda_\xi}$ exist. $l_{t\nu}^\Lambda $ exists iff
$\nu(x_t|x_{<t})$ exists iff $\nu(x_{<t})>0$ iff
$w_\nu(x_{<t})>0$.
\beqn
  l_{t\xi}^{\Lambda} \;=\;
  \sum_\nu w_\nu(x_{<t}) l_{t\nu}^\Lambda  \;<\;
  \sum_\nu w_\nu(x_{<t}) l_{t\nu}^{\Lambda_\xi} \;=\;
  l_{t\xi}^{\Lambda_\xi} \;\leq\;
  l_{t\xi}^{\Lambda}
\eeqn
The two equalities follow from inserting (\ref{xirel}) into
(\ref{mxexloss}).
The strict inequality follows from the assumption and
$w_\nu(x_{<t})>0$. The last inequality follows from the fact that
$\Lambda_\xi$ minimizes by definition (\ref{xlrdef}) the
$\xi$-expected loss (similarly to (\ref{Lmuopt})). The
contradiction $l_{t\xi}^{\Lambda} < l_{t\xi}^{\Lambda}$ proves
Pareto-optimality of $\xi$ w.r.t.\ $l_t$.

In the same way we can prove Pareto-optimality of $\xi$ w.r.t.\
the total loss $L_n$ by defining the $\rho$-expected
total losses
\beqn\label{mxexloss2}
  L_{n\rho}^{\Lambda}\;:=\;\sum_{t=1}^n\sum_{x_{<t}}\rho(x_{<t})l_{t\rho}^\Lambda (x_{<t})
  \;=\;\sum_{t=1}^n\sum_{x_{1:t}}\rho(x_{1:t})\ell_{x_ty_t^\Lambda}
\eeqn
for a predictor $\Lambda$, and by assuming
$L_{n\nu}^{\Lambda} \leq L_{n\nu}^{\Lambda_\xi}$ for all $\nu$ and strict
inequality for some $\nu$, from which we get the contradiction
$L_{n\xi}^{\Lambda}=\sum_\nu w_\nu L_{n\nu}^{\Lambda}<\sum_\nu
w_\nu L_{n\nu}^{\Lambda_\xi}=L_{n\xi}^{\Lambda_\xi}\leq
L_{n\xi}^{\Lambda}$ with the help of (\ref{xidefsp}).
The instantaneous and total expected errors $e_t$ and $E_n$ can be
considered as special loss functions.

Pareto-optimality of $\xi$ w.r.t.\ $s_t$ (and hence $S_n$) can be
understood from geometrical insight. A formal proof for $s_t$ goes
as follows: With the abbreviations $i =x_t$,
$y_{\nu i} = \nu(x_t|x_{<t})$, $z_i = \xi(x_t|x_{<t})$,
$r_i = \rho(x_t|x_{<t})$, and $w_\nu = w_\nu(x_{<t}) \geq 0$ we
ask for a vector $\v r$ with
$\sum_i(y_{\nu i} - r_i)^2\leq\sum_i(y_{\nu i} - z_i)^2\;\forall\nu$.
This implies
\bqan
  0 &\geq&
  \sum_\nu w_\nu\Big[\sum_i(y_{\nu i}\!-\!r_i)^2-\sum_i(y_{\nu i}\!-\!z_i)^2\Big]
  \;=\; \sum_\nu w_\nu\Big[\sum_i -2y_{\nu i}r_i+r_i^2+2y_{\nu i}z_i-z_i^2\Big]
\\
  &=& \sum_i -2z_ir_i+r_i^2+2z_iz_i-z_i^2
  \;=\; \sum_i (r_i\!-\!z_i)^2 \;\geq\; 0
\eqan
where we have used $\sum_\nu w_\nu = 1$ and $\sum_\nu
w_\nu y_{\nu i} = z_i$ (\ref{xirel}). $0\geq\sum_i(r_i-z_i)^2\geq 0$
implies $\v r = \v z$ proving unique Pareto-optimality of
$\xi$ w.r.t.\ $s_t$. Similarly for $d_t$ the assumption $\sum_i
y_{\nu i}\ln{y_{\nu i}\over r_i} \leq \sum_i y_{\nu i}\ln{y_{\nu i}\over
z_i}\;\forall\nu$ implies
\beqn
  0 \;\geq\;
  \sum_\nu w_\nu \Big[\sum_i y_{\nu i}\ln{y_{\nu i}\over r_i} - y_{\nu i}\ln{y_{\nu i}\over z_i}\Big]
  \;=\; \sum_\nu w_\nu \sum_i y_{\nu i}\ln{z_i\over r_i}
  \;=\; \sum_i z_i\ln{z_i\over r_i} \;\geq\; 0
\eeqn
which implies $\v r = \v z$ proving unique Pareto-optimality
of $\xi$ w.r.t.\ $d_t$. The proofs for $S_n$ and $D_n$ are similar.
\qed

We have proven that $\xi$ is {\em uniquely} Pareto-optimal w.r.t.\
$s_t$, $S_n$, $d_t$ and $D_n$. In the case of $e_t$, $E_n$, $l_t$
and $L_n$ there are other $\rho \neq \xi$ with $\F(\nu,\rho) =
\F(\nu,\xi)\forall \nu$, but the actions/predictions they invoke
are unique ($y_t^{\Lambda_\rho} = y_t^{\Lambda_\xi})$ (if ties in
$\arg\max_{y_t}$ are broken in a consistent way), and this is all
that counts.

Note that $\xi$ is {\em not} Pareto-optimal w.r.t.\ to {\em all}
performance measures. Counterexamples can be given for
$\F(\nu,\xi)=\sum_{x_t}|\nu(x_t|x_{<t})-\xi(x_t|x_{<t})|^\alpha$
for $\alpha\neq 2$, e.g.\ $a_t$ and $V_n$. Nevertheless, for all
performance measures which are relevant from a decision theoretic
point of view, i.e.\ for all loss functions $l_t$ and $L_n$, $\xi$
has the welcome property of being Pareto-optimal.

\indxs{balanced}{Pareto-optimality}
Pareto-optimality should be regarded as a necessary condition for
a prediction scheme aiming to be optimal. From a practical point
of view a significant decrease of $\F$ for many $\nu$ may be
desirable even if this causes a small increase of $\F$ for a few
other $\nu$. One can show that such a ``balanced''
improvement is (not) possible in the following sense:
For instance, by using $\tilde\Lambda$ instead of $\Lambda_\xi$,
the $w_\nu$-expected loss may increase or decrease, i.e.\
$L_{n\nu}^{\tilde\Lambda} \lessgtr L_{n\nu}^{\Lambda_\xi}$, but
on average, the loss can not decrease, since $\sum_\nu
w_\nu[L_{n\nu}^{\tilde \Lambda} -
L_{n\nu}^{\Lambda_\xi}]=L_{n\xi}^{\tilde \Lambda} -
L_{n\xi}^{\Lambda_\xi}\geq 0$, where we have used linearity of
$L_{n\rho}$ in $\rho$ and $L_{n\xi}^{\Lambda_\xi} \leq
L_{n\xi}^\Lambda$.
In particular, a loss increase by an amount $\Delta_\lambda$ in
only a single environment $\lambda$, can cause a decrease by at
most the same amount times a factor ${w_\lambda\over w_\eta}$ in
some other environment $\eta$, i.e.\ a loss increase can only
cause a smaller decrease in simpler environments, but a scaled
decrease in more complex environments.
%
\index{no free lunch}\index{free lunch}\index{Occam's razor}
\indxs{environment}{random}
We do not regard this as a ``No Free Lunch''
(NFL) theorem \cite{Wolpert:97}. Since most environments are
completely random, a small concession on the loss in each of these
completely uninteresting environments provides enough margin
to yield distinguished performance on the few
non-random (interesting) environments. Indeed, we would interpret
the NFL theorems for optimization and search \citein{Wolpert:97}
as balanced Pareto-optimality results. Interestingly, whereas for
prediction only Bayes-mixes are Pareto-optimal, for search and
optimization every algorithm is Pareto-optimal.

The term {\em Pareto-optimal} has been taken from the
economics literature, but there is the closely related notion of
unimprovable strategies \cite{Borovkov:98} or admissible
estimators \cite{Ferguson:67} in statistics for parameter
estimation, for which results similar to Theorem \ref{thPareto}
exist. Furthermore, it would be interesting to show under which
conditions, the class of {\em all} Bayes-mixtures (i.e.\ with all
possible values for the weights) is complete in the sense that
{\em every} Pareto-optimal strategy can be based on a Bayes-mixture.
Pareto-optimality is sort of a minimal demand on a prediction
scheme aiming to be optimal. A scheme which is not even
Pareto-optimal cannot be regarded as optimal in any reasonable
sense. Pareto-optimality of $\xi$ w.r.t.\ most performance
measures emphasizes the distinctiveness of Bayes-mixture
strategies.

\subsection{On the Optimal Choice of Weights}\label{subsecWeights}
\index{weights!optimal} In the following we indicate the
dependency of $\xi$ on $w$ explicitly by writing $\xi_w$. We have
shown that the $\Lambda_{\xi_w}$ prediction schemes are (balanced)
Pareto-optimal, i.e.\ that {\em no} prediction scheme $\Lambda$
(whether based on a Bayes mix or not) can be uniformly better.
Least assumptions on the environment are made for $\M$ which are
as large as possible. In Section \ref{secPCM} we have discussed
the set $\M$ of all enumerable semimeasures which we regarded as
sufficiently large from a computational point of view (see
\citealt{Schmidhuber:02gtm,Hutter:03unipriors} for even larger sets,
but which are still in the computational realm). Agreeing on this
$\M$ still leaves open the question of how to choose the weights
(prior beliefs) $w_\nu$, since every $\xi_w$ with
$w_\nu>0$ $\forall\nu$ is Pareto-optimal and leads asymptotically to
optimal predictions.

We have derived bounds for the mean squared sum
$S_{n\nu}^{\xi_w}\leq\ln w_\nu^{-1}$ and for the loss regret
$L_{n\nu}^{\Lambda_{\xi_w}}-L_{n\nu}^{\Lambda_\nu}\leq 2\,\ln
w_\nu^{-1}+2\sqrt{\ln w_\nu^{-1}L_{n\nu}^{\Lambda_\nu}}$.
All bounds monotonically decrease with increasing $w_\nu$. So it
is desirable to assign high weights to all $\nu\in\M$. Due to the
(semi)probability constraint $\sum_\nu w_\nu\leq 1$ one has to
find a compromise.%
\footnote{All results in this paper have been stated and proven
for probability measures $\mu$, $\xi$ and $w_\nu$, i.e.\
$\sum_{x_{1:t}}\xi(x_{1:t}) = \sum_{x_{1:t}}\mu(x_{1:t}) =
\sum_\nu w_\nu = 1$. On the other hand, the class $\M$ considered
here is the class of all \indxs{enumerable}{semimeasure}
enumerable semimeasures and $\sum_\nu
w_\nu<1$. In general, each of the following 4 items could be semi
($<$) or not ($=$): ($\xi$, $\mu$, $\M$, $w_\nu$), where $\M$ is
semi if some elements are semi. Six out of the $2^4$ combinations
make sense. Convergence (Theorem \ref{thConv}), the error bound (Theorem
\ref{thErrBnd}), the loss bound (\ref{th3}), as well as most other
statements hold for $(<,=,<,<)$, but not for $(<,<,<,<)$.
Nevertheless, $\xi\to\mu$ holds also for $(<,<,<,<)$ with maximal
$\mu$ \idx{semi-probability}, i.e.\ fails with
$\mu$ semi-probability 0.}%
\indxs{enumerable}{weights}
In the following we will argue that in the class of enumerable
weight functions with short program there is an optimal
compromise, namely $w_\nu=2^{-K(\nu)}$.

Consider the class of enumerable weight functions with short
programs, namely ${\cal V}:=\{v_{(.)}:\M\to \Set{R}^+$ with
$\sum_\nu v_\nu\leq 1$ and $K(v)=O(1)\}$. Let $w_\nu:=2^{-K(\nu)}$
and $v_{(\cdot)}\in\cal V$.
Corollary 4.3.1 of \tcite[p255]{Li:97} says that $K(x)\leq -\log_2
P(x) + K(P) + O(1)$ for all $x$ if $P$ is an enumerable discrete
semimeasure. Identifying $P$ with $v$ and $x$ with (the program
index describing) $\nu$ we get
\beqn
  \ln w_\nu^{-1} \;\leq \ln v_\nu^{-1} + O(1).
\eeqn
This means that the bounds for $\xi_w$ depending on $\ln
w_\nu^{-1}$ are at most $O(1)$ larger than the bounds for $\xi_v$
depending on $\ln v_\nu^{-1}$. So we lose at most an additive
constant of order one in the bounds when using $\xi_w$ instead of
$\xi_v$. In using $\xi_w$ we are on the safe
side, getting (within $O(1)$) best bounds for {\em all}
environments.

\ftheorem{thSolOpt}{Optimality of universal weights}{
\indxs{optimal}{weights}
Within the set $\cal V$ of enumerable weight functions with short
program, the universal weights $w_\nu=2^{-K(\nu)}$ lead to the
smallest loss bounds within an additive (to $\ln w_\mu^{-1}$)
constant in all enumerable environments.
}

\noindent Since the above justifies the use of $\xi_w$, and
$\xi_w$ assigns high probability to an environment if and only if
it has low (Kolmogorov) complexity, one may interpret the result
as a justification of Occam's razor.$\!$\footnote{The {\em only
if} direction can be shown by a more easy and direct
argument \cite{Schmidhuber:02gtm}.} But
note that this is more of a \idx{bootstrap} argument, since we
implicitly used Occam's razor to justify the restriction to
enumerable semimeasures. We also considered only weight functions
$v$ with low complexity $K(v)=O(1)$. What did not enter as an
assumption but came out as a result is that the specific universal
weights $w_\nu=2^{-K(\nu)}$ are optimal.

On the other hand, this choice for $w_\nu$ is not unique (even not
within a constant factor). For instance, for $0<v_\nu=O(1)$ for
$\nu=\xi_w$ and $v_\nu$ arbitrary (e.g.\ 0) for all other $\nu$,
the obvious dominance $\xi_\nu\geq v_\nu\nu$ can be improved to
$\xi_\nu\geq c\cdot w_\nu\nu$, where $0<c=O(1)$ is a universal
constant. Indeed, formally every choice of weights
$v_\nu>0\,\forall\nu$ leads within a multiplicative constant to
the same universal distribution, but this constant is not
necessarily of ``acceptable'' size. Details will be presented
elsewhere.

\section{Miscellaneous}\label{secMisc}

This section discusses miscellaneous topics.
Section \ref{secCPC} generalizes the setup to continuous
probability classes $\M = \{\mu_\theta\}$ consisting of continuously
parameterized distributions $\mu_\theta$ with parameter
$\theta \in \Set{R}^d$. Under certain smoothness and regularity
conditions a bound for the relative entropy between $\mu$ and
$\xi$, which is central for all presented results, can still be
derived. The bound depends on the Fisher information of $\mu$ and
grows only logarithmically with $n$, the intuitive reason being
the necessity to describe $\theta$ to an accuracy $O(n^{-1/2})$.
Section \ref{secFApp} describes two ways of using the prediction
schemes for partial sequence prediction, where not every symbol
needs to be predicted. Performing and predicting a sequence of
independent experiments and online learning of classification
tasks are special cases.
In Section \ref{secWM} we compare the universal prediction scheme
studied here to the popular predictors based on expert advice
(PEA)
\cite{Littlestone:89,Vovk:92,Littlestone:94,Cesa:97,Haussler:98,Kivinen:99}.
Although the algorithms, the settings, and the proofs are quite
different, the PEA bounds and our error bound have the same
structure.
Finally, in Section \ref{secOut} we outline possible extensions of
the presented theory and results, including infinite alphabets,
delayed and probabilistic prediction, active systems influencing
the environment, learning aspects, and a unification with PEA.

\subsection{Continuous Probability Classes $\M$}\label{secCPC}
\indxs{probability class}{continuous}
\indxs{hypothesis class}{continuous}
\indxs{Bernoulli}{process}
\indxs{i.i.d.}{process}
We have considered thus far countable probability classes
$\M$, which makes sense from a computational point of view as
emphasized in Section \ref{secPCM}. On the other hand in
statistical parameter estimation one often has a continuous
hypothesis class (e.g.\ a Bernoulli($\theta$) process with unknown
$\theta \in[0,1]$). Let
\beqn
  \M \;:=\; \{\mu_\theta:\theta\in\Theta\subseteq \Set{R}^d\}
\eeqn
be a family of probability distributions parameterized by a
$d$-dimensional continuous parameter $\theta$. Let $\mu \equiv
\mu_{\theta_0} \in \M$ be the true generating distribution and
$\theta_0$ be in the interior of the compact set $\Theta$. We may
restrict $\M$ to a countable dense subset, like $\{\mu_\theta\}$
with computable (or rational) $\theta$. If $\theta_0$ is itself a
computable real (or rational) vector then Theorem \ref{thConv} and bound
(\ref{thULoss}) apply. From a practical point of view the assumption
of a computable $\theta_0$ is not so serious. It is more from a
traditional analysis point of view that one would like quantities
and results depending smoothly on $\theta$ and not in a weird
fashion depending on the computational complexity of $\theta$.
\indxs{weights}{continuous} For instance, the weight $w(\theta)$
is often a continuous probability density
\beq\label{xidefc}
  \xi(x_{1:n}) \;:=\; \int_\Theta \! d\theta\,
  w(\theta)\!\cdot\!\mu_\theta(x_{1:n}), \qquad
  \int_\Theta \! d\theta\,
  w(\theta) \;=\; 1, \qquad
  w(\theta) \;\geq\; 0.
\eeq
%
\noindent The most important property of $\xi$ used in this work was
$\xi(x_{1:n})  \geq  w_\nu \cdot \nu(x_{1:n})$
which has been obtained from (\ref{xidefsp}) by dropping the sum
over $\nu$. The analogous construction here is to restrict the
integral over $\Theta$ to a small vicinity $N_\delta$ of $\theta$.
For sufficiently smooth $\mu_\theta$ and $w(\theta)$ we expect
$\xi(x_{1:n}) \approxgeq |N_{\delta_n}|  \cdot
w(\theta)  \cdot  \mu_\theta(x_{1:n})$, where
$|N_{\delta_n}|$ is the volume of $N_{\delta_n}$. This in turn
leads to $D_n \approxleq
\ln{w_\mu^{-1}}+\ln|N_{\delta_n}|^{-1}$, where
$w_\mu:=w(\theta_0)$. $N_{\delta_n}$ should be the largest possible
region in which $\ln\mu_\theta$ is approximately flat on average.
The averaged instantaneous, mean, and total curvature
matrices of $\ln\mu$ are\index{curvature matrix}
\bqan\label{fisher}
  j_t(x_{<t}) &:=& \E_t
  \nabla_\theta\ln\mu_\theta(x_t|x_{<t})
  \nabla^T_\theta\ln\mu_\theta(x_t|x_{<t})_{|\theta=\theta_0}
  ,\qquad \bar\jmath_n \;:=\; {\textstyle{1\over n}}J_n
  \\
  J_n &:=& \sum_{t=1}^n \E_{<t} j_t(x_{<t}) \;=\;
  \E_{1:n} \nabla_\theta\ln\mu_\theta(x_{1:n})
  \nabla^T_\theta\ln\mu_\theta(x_{1:n})_{|\theta=\theta_0}
  \nonumber
\eqan
\index{Fisher information}\index{information!Fisher}%
\indxs{parametric}{complexity}%
They are the Fisher information of $\mu$ and may be viewed as
measures of the parametric complexity of $\mu_\theta$ at
$\theta=\theta_0$. The last equality can be shown by using the
fact that the $\mu$-expected value of $\nabla\ln\mu \cdot
\nabla^T\ln\mu$ coincides with $-\nabla\nabla^T\ln\mu$ (since $\X$
is finite) and a similar equality as in (\ref{entropy})
for $D_n$.

\ftheorem{thCEB}{Continuous Entropy Bound}{
\indxs{entropy}{bound}\index{continuous!entropy bound}
\indxs{relative}{entropy}
Let $\mu_\theta$ be twice continuously differentiable at $\theta_0
\in \Theta \subseteq \Set{R}^d$ and $w(\theta)$ be continuous and
positive at $\theta_0$. Furthermore we assume that the inverse of
the mean Fisher information matrix $(\bar\jmath_n)^{-1}$ exists,
is bounded for $n\to\infty$, and is uniformly (in $n$) continuous at
$\theta_0$. Then the relative entropy $D_n$ between $\mu \equiv
\mu_{\theta_0}$ and $\xi$ (defined in (\ref{xidefc})) can be
bounded by
\beqn
  D_n \;:=\;
  \E_{1:n}\textstyle
  \ln{\mu(x_{1:n}) \over \xi(x_{1:n})} \;\;\leq\;\;
  \ln{w_\mu^{-1}} + {d\over 2}\ln{n\over 2\pi} +
  {1\over 2}\ln\det\bar\jmath_n + o(1)
  \;=:\; b_\mu
\eeqn
where $w_\mu\equiv w(\theta_0)$ is the weight density (\ref{xidefc}) of
$\mu$ in $\xi$ and $o(1)$ tends to zero for $n \to\infty$.
}

\paradot{Proof sketch}
\noindent For independent and identically distributed
distributions $\mu_\theta(x_{1:n}) = \mu_\theta(x_1) \cdot...\cdot
\mu_\theta(x_n)\,\forall\theta$ this bound has been proven
in \cite[Theorem 2.3]{Clarke:90}. In this case $J^{[CB90]}(\theta_0) \equiv
\bar\jmath_n \equiv j_n$ independent of $n$. For stationary
($k^{th}$-order) Markov processes $\bar\jmath_n$ is also constant.
The proof generalizes to arbitrary $\mu_\theta$ by replacing
$J^{[CB90]}(\theta_0)$ with $\bar\jmath_n$ everywhere in their
proof. For the proof to go through, the vicinity $N_{\delta_n} :=
\{\theta: ||\theta-\theta_0||_{\bar\jmath_n} \leq \delta_n\}$ of
$\theta_0$ must contract to a point set $\{\theta_0\}$ for
$n\to\infty$ and $\delta_n \to 0$. $\bar\jmath_n$ is always
positive semi-definite as can be seen from the definition. The
boundedness condition of $\bar\jmath_n^{-1}$ implies a strictly
positive lower bound independent of $n$ on the eigenvalues of
$\bar\jmath_n$ for all sufficiently large $n$, which ensures
$N_{\delta_n}\to\{\theta_0\}$. The uniform continuity of
$\bar\jmath_n$ ensures that the remainder $o(1)$ from the Taylor
expansion of $D_n$ is independent of $n$. Note that twice
continuous differentiability of $D_n$ at $\theta_0$
\cite[Condition 2]{Clarke:90} follows for finite $\X$ from twice
continuous differentiability of $\mu_\theta$. Under some
additional technical conditions one can even prove an equality $
  D_n =
  \ln{w_\mu^{-1}} + {d\over 2}\ln{n\over 2\pi e} +
  {1\over 2}\ln\det\bar\jmath_n + o(1)
$
for the i.i.d.\ case \cite[(1.4)]{Clarke:90}, which is probably
also valid for general $\mu$. \qed

The $\ln{w_\mu^{-1}}$ part in the bound is the same as for
countable $\M$. The ${d\over 2}\ln{n\over 2\pi}$ can be understood
as follows: Consider $\theta \in [0,1)$ and restrict the
continuous $\M$ to $\theta$ which are finite binary fractions.
Assign a weight $w(\theta) \approx 2^{-l}$ to a $\theta$ with
binary representation of length $l$.
$D_n\approxleq l \cdot \ln 2$ in this case. But what if
$\theta$ is not a finite binary fraction? A continuous parameter
can typically be estimated with accuracy $O(n^{-1/2})$ after $n$
observations.
\indxs{parameter}{estimate}
The data do not allow to distinguish a
$\tilde\theta$ from the true $\theta$ if
$|\tilde\theta - \theta|<O(n^{-1/2})$. There is such a
$\tilde\theta$ with binary representation of length $l = \log_2
O(\sqrt n)$. Hence we expect $D_n\approxleq \odt\ln n+O(1)$ or
${d\over 2}\ln n+O(1)$ for a $d$-dimensional parameter space. In
general, the $O(1)$ term depends on the parametric complexity of
$\mu_\theta$ and is explicated by the third ${1\over
2}\ln\det\bar\jmath_n$ term in Theorem \ref{thCEB}. See
\tcite[p454]{Clarke:90} for an alternative explanation. Note that a
uniform weight $w(\theta)={1\over|\Theta|}$ does not lead to a
uniform bound unlike the discrete case. A uniform bound is
obtained for Bernando's (or in the scalar case Jeffreys')
reference prior
$w(\theta) \sim \sqrt{\det\bar\jmath_\infty(\theta)}$ if
$\jmath_\infty$ exists \cite{Rissanen:96}.
\index{Jeffreys' prior}\index{prior!Jeffreys}
\index{Bernando's prior}\index{prior!Bernando}

For a finite alphabet $\X$ we consider throughout the paper,
$j_t^{-1}<\infty$ independent of $t$ and $x_{<t}$ in case of
i.i.d.\ sequences. More generally, the conditions of Theorem
\ref{thCEB} are satisfied for the practically very important class
of stationary ($k$-th order) finite-state Markov processes ($k=0$
is i.i.d.).

Theorem \ref{thCEB} shows that Theorems \ref{thConv} and
\ref{thErrBnd} are also applicable to the case of continuously
parameterized probability classes. Theorem \ref{thCEB} is also
valid for a mixture of the discrete and continuous cases $\xi  =
\sum_a\int d\theta\, w^a(\theta)\,\mu^a_\theta$ with $\sum_a\int
d\theta\, w^a(\theta)  =  1$.

\subsection{Further Applications}\label{secFApp}

\paradot{Partial sequence prediction}
\index{application!partial sequence prediction}
There are (at least) two ways to treat partial
sequence prediction. With this we mean that not every symbol of the
sequence needs to be predicted, say given sequences of the
form $z_1x_1...z_nx_n$ we want to predict the $x's$ only.
The first way is to keep the $\Lambda_\rho$ prediction schemes of the last
sections mainly as they are, and use a time dependent loss
function, which assigns zero loss $\ell^t_{zy} \equiv 0$ at the
$z$ positions. Any dummy prediction $y$ is then consistent with
(\ref{xlrdef}). The losses for predicting $x$ are generally
non-zero.
This solution is satisfactory as long as the $z's$ are drawn from
a probability distribution. The second (preferable) way does not
rely on a probability distribution over the $z$. We replace all
distributions $\rho(x_{1:n})$ ($\rho=$ $\mu$, $\nu$, $\xi$)
everywhere by distributions $\rho(x_{1:n}|z_{1:n})$
conditioned on $z_{1:n}$. The $z_{1:n}$ conditions cause nowhere
problems as they can essentially be thought of as fixed (or as
oracles or spectators). So the bounds in Theorems
\ref{thConv}...\ref{thCEB} also hold in this case for all
individual $z$'s.

\paradot{Independent experiments and classification}
\index{classification}%
\index{application!classification}%
\index{application!i.i.d. experiments}%
\index{experiments!i.i.d.}%
A typical experimental situation is a sequence of independent
(i.i.d) experiments, predictions and observations. At time $t$ one
arranges an experiment $z_t$ (or observes data $z_t$), then tries
to make a prediction, and finally observes the true outcome $x_t$.
Often one has a parameterized class of models (hypothesis space)
$\mu_\theta(x_t|z_t)$ and wants to infer the true $\theta$ in
order to make improved predictions. This is a special case of
partial sequence prediction, where the hypothesis space
$\M=\{\mu_\theta(x_{1:n}|z_{1:n})=\mu_\theta(x_1|z_1)\cdot...\cdot\mu_\theta(x_n|z_n)\}$ consists of
i.i.d.\ distributions, but note that $\xi$ is not i.i.d. This is
the same setting as for on-line learning of classification tasks,
where a $z \in \cal Z$ should be classified as an $x \in \X$.

\subsection{Prediction with Expert Advice}\label{secWM}
\index{weighted majority algorithm}\index{algorithm!weighted majority}
\index{prediction!with expert advice}
\index{learning!with expert advice}
\index{aggregating strategy}\index{stragegy!aggregating}
\index{boosting}
\index{hedge algorithm}\index{algorithm!hegde}%
There are two schools of universal sequence prediction: We
considered expected performance bounds for Bayesian prediction
based on mixtures of environments, as is common in information
theory and statistics \cite{Merhav:98}. The other approach are
predictors based on expert advice (PEA) with worst case loss
bounds in the spirit of Littlestone, Warmuth, Vovk and others. We
briefly describe PEA and compare both approaches. For a more
comprehensive comparison see \tcite{Merhav:98}. In the following we
focus on topics not covered \citein{Merhav:98}.
PEA was invented \citein{Littlestone:89,Littlestone:94} and
\tcite{Vovk:92} and further developed
\citein{Cesa:97,Haussler:98,Kivinen:99} and by many others. Many
variations known by many names (prediction/learning with expert
advice, weighted majority/average, aggregating strategy,
hedge algorithm, ...) have meanwhile been invented. Early works in
this direction are \tcite{Dawid:84,Rissanen:89}. See \tcite{Vovk:99}
for a review and further references. We describe the setting and
basic idea of PEA for binary alphabet. Consider a finite binary
sequence $x_1x_2...x_n \in \{0,1\}^n$ and a finite set $\cal E$ of
experts $e \in \cal E$ making predictions $x_t^e$ in the unit
interval $[0,1]$ based on past observations $x_1x_2...x_{t-1}$.
The loss of expert $e$ in step $t$ is defined as $|x_t - x_t^e|$.
In the case of binary predictions $x_t^e \in \{0,1\}$, $|x_t -
x_t^e|$ coincides with our error measure (\ref{rhoerr}). The PEA
algorithm $p_{\beta n}$ combines the predictions of all
experts.\indxs{prediction}{combining experts} It forms its own
prediction\footnote{The original PEA version \cite{Littlestone:89}
had discrete prediction $x_t^p \in \{0,1\}$ with (necessarily)
twice as many errors as the best expert and is only of historical
interest any more.} $x_t^p \in [0,1]$ according to some weighted
average of the expert's predictions $x_t^e$. \index{update
weights}\index{weights!update rule} There are certain update rules
for the weights depending on some parameter $\beta$. Various
bounds for the total loss $L_p(\v x) := \sum_{t=1}^n|x_t -
x_t^p|$ of PEA in terms of the total loss $L_\eps(\v x) :=
\sum_{t=1}^n|x_t - x_t^\eps|$ of the best expert $\eps \in \cal E$
have been proven. It is possible to fine tune $\beta$ and to
eliminate the necessity of knowing $n$ in advance. The first bound
of this kind has been obtained \citein{Cesa:97}:
\beq\label{wmbnd}
  L_p(\v x) \;\leq\; L_\eps(\v x)+2.8\ln|{\cal
  E}|+4\sqrt{L_\eps(\v x)\ln|{\cal E}|}.
\eeq
The constants 2.8 and 4 have been improved
\citein{Auer:00,Yaroshinsky:01}. The last
bound in Theorem \ref{thErrBnd} with $S_n\leq D_n \leq \ln|\M|$ for
uniform weights and with $E_n^{\Theta_\mu}$ increased to
$E_n^\Theta$ reads
\beqn
  E_n^{\Theta_\xi} \;\leq\; E_n^\Theta + 2\ln|{\M}| +
  2\sqrt{E_n^\Theta\ln|{\M}|}.
\eeqn
\indxs{structure}{loss bound} It has a quite similar structure as
(\ref{wmbnd}), although the algorithms, the settings, the proofs,
and the interpretation are quite different. Whereas PEA performs
well in any environment, but only relative to a given set of
experts $\cal E$, our $\Theta_\xi$ predictor competes with the
best possible $\Theta_\mu$ predictor (and hence with any other
$\Theta$ predictor), but only in expectation and for a given set
of environments $\M$. PEA depends on the set of experts,
$\Theta_\xi$ depends on the set of environments $\M$.
The basic $p_{\beta n}$ algorithm has been extended in different
directions: incorporation of different initial weights ($|{\cal
E}|\leadsto w_\nu^{-1}$) \cite{Littlestone:89,Vovk:92},
more general loss functions \cite{Haussler:98}, continuous valued
outcomes \cite{Haussler:98}, and multi-dimensional predictions
\cite{Kivinen:99} (but not yet for the absolute loss). The work
of \tcite{Yamanishi:98} lies somewhat in
between PEA and this work; ``PEA'' techniques are used to prove
expected loss bounds (but only for sequences of independent
symbols/experiments and limited classes of loss functions).
\indxs{probabilistic}{forecast}%
\indxs{continuous}{forecast}%
Finally, note that the predictions of PEA are continuous. This is
appropriate for weather forecasters which announce the probability
of rain, but the {\em decision} to wear sunglasses or to take an
umbrella is binary, and the suffered loss depends on this binary
decision, and not on the probability estimate. It is possible to
convert the continuous prediction of PEA into a probabilistic
binary prediction by predicting 1 with probability $x_t^p \in
[0,1]$. $|x_t - x_t^p|$ is then the probability of making an
error. Note that the expectation is taken over the probabilistic
prediction, whereas for the deterministic $\Theta_\xi$ algorithm
the expectation is taken over the environmental distribution
$\mu$. The multi-dimensional case \cite{Kivinen:99} could then be
interpreted as a (probabilistic) prediction of symbols over an
alphabet $\X = \{0,1\}^d$, but error bounds for the absolute loss
have yet to be proven. In \tcite{Freund:97} the regret is bounded
by $\ln|{\cal E}|+\sqrt{2\tilde L\,\ln|{\cal E}|}$ for arbitrary
unit loss function and alphabet, where $\tilde L$ is an upper
bound on $L_\eps$, which has to be known in advance. It would be
interesting to generalize PEA and bound (\ref{wmbnd}) to arbitrary
alphabet and weights and to general loss functions with
probabilistic interpretation.

\subsection{Outlook}\label{secOut}
In the following we discuss several directions in which the
findings of this work may be extended.

\paradot{Infinite alphabet}
\indxs{alphabet}{infinite}%
\indxs{alphabet}{countable}%
\indxs{alphabet}{continuous}%
In many cases the basic prediction unit is not a letter, but a
number (for inducing number sequences), or a word (for completing
sentences), or a real number or vector (for physical
measurements). The prediction may either be generalized to a block
by block prediction of symbols or, more suitably, the finite
alphabet $\X$ could be generalized to countable (numbers,
words) or continuous (real or vector) alphabets. The presented theorems
are independent of the size of $\X$ and hence should
generalize to countably infinite alphabets by appropriately taking
the limit $|\X| \to \infty$ and to continuous alphabets by
a denseness or separability argument. Since the proofs are also
independent of the size of $\X$ we may directly replace all
finite sums over $\X$ by infinite sums or integrals and
carefully check the validity of each operation. We expect all
theorems to remain valid in full generality, except for minor
technical existence and convergence constraints.

\indxs{infinite}{prediction space}
An infinite prediction space $\Y$ was no problem at all as long
as we assumed the existence of $y_t^{\Lambda_\rho} \in \Y$
(\ref{xlrdef}). In case $y_t^{\Lambda_\rho} \in \Y$ does not
exist one may define $y_t^{\Lambda_\rho} \in \Y$ in a way to
achieve a loss at most $\eps_t = o(t^{-1})$ larger than the
infimum loss. We expect a small finite correction of the order of
$\eps=\sum_{t=1}^\infty\eps_t<\infty$ in the loss bounds somehow.

\paradot{Delayed \& probabilistic prediction}
\indxs{delayed}{prediction}\indxs{probabilistic}{prediction}%
The $\Lambda_\rho$ schemes and theorems may be generalized to
delayed sequence prediction, where the true symbol $x_t$ is given
only in cycle $t + d$. A delayed feedback is common in many
practical problems. We expect bounds with $D_n$ replaced by
$d \cdot D_n$. Further, the error bounds for the
probabilistic suboptimal $\xi$ scheme defined and analyzed
\citein{Hutter:99errbnd} can also be generalized to arbitrary alphabet.

\paradot{More active systems}
\indxs{active}{system}%
\indxs{strategy}{greedy}%
\indxs{influence}{environment}%
Prediction means guessing the future, but not influencing it. A
small step in the direction of more active systems was to allow
the $\Lambda$ system to act and to receive a loss $\ell_{x_t y_t}$
depending on the action $y_t$ and the outcome $x_t$. The
probability $\mu$ is still independent of the action, and the loss
function $\ell^t$ has to be known in advance. This ensures that
the greedy strategy (\ref{xlrdef}) is optimal. The loss function
may be generalized to depend not only on the history $x_{<t}$, but
also on the historic actions $y_{<t}$ with $\mu$ still independent
of the action. It would be interesting to know whether the scheme
$\Lambda$ and/or the loss bounds generalize to this case. The full
model of an acting agent influencing the environment has been
developed \citein{Hutter:01aixi}. Pareto-optimality and
asymptotic bounds are proven \citein{Hutter:02selfopt}, but a lot
remains to be done in the active case.

\paradot{Miscellaneous}
\index{learning}\indxs{Maximum-Likelihood}{prediction}%
Another direction is to investigate the learning aspect of
universal prediction. Many prediction schemes explicitly learn and
exploit a model of the environment. Learning and exploitation are
melted together in the framework of universal Bayesian prediction.
A separation of these two aspects in the spirit of hypothesis
learning with MDL \cite{Vitanyi:00} could lead to new insights.
Also, the separation of noise from useful data, usually an
important issue \cite{Gacs:01}, did not play a role here. The
attempt at an information theoretic interpretation of Theorem
\ref{thWin} may be made more rigorous in this or another way. In
the end, this may lead to a simpler proof of Theorem \ref{thWin}
and maybe even for the loss bounds. A unified picture of the loss
bounds obtained here and the loss bounds for predictors based on
expert advice (PEA) could also be fruitful. Yamanishi
\citey{Yamanishi:98} used PEA methods to prove expected loss
bounds for Bayesian prediction, so maybe the proof technique
presented here could be used {\em vice versa} to prove more
general loss bounds for PEA. Maximum-likelihood or MDL predictors
may also be studied. For instance, $2^{-K(x)}$ (or some of its
variants) is a close approximation of $\xi_U$, so one may think
that predictions based on (variants of) $K$ may be as good as
predictions based on $\xi_U$, but it is easy to see that $K$
completely fails for predictive purposes. Also, more promising
variants like the monotone complexity $\Km$ and universal two-part
MDL, both extremely close to $\xi_U$, fail in certain situations
\cite{Hutter:03unimdl}. Finally, the system should be applied to
specific induction problems for specific $\M$ with computable
$\xi$.

\section{Summary}\label{secSPConc}

We compared universal predictions based on Bayes-mixtures $\xi$ to
the infeasible informed predictor based on the unknown true
generating distribution $\mu$. Our main focus was on a
decision-theoretic setting, where each prediction $y_t\in\X$ (or
more generally action $y_t\in\Y$) results in a loss $\ell_{x_t
y_t}$ if $x_t$ is the true next symbol of the sequence. We have
shown that the $\Lambda_\xi$ predictor suffers only slightly more
loss than the $\Lambda_\mu$ predictor.
We have shown that the derived error and loss bounds cannot be
improved in general, i.e.\ without making extra assumptions on
$\ell$, $\mu$, $\M$, or $w_\nu$. Within a factor of 2 this is also
true for any $\mu$ independent predictor. We have also shown
Pareto-optimality of $\xi$ in the sense that there is no other
predictor which performs at least as well in all environments
$\nu\in\M$ and strictly better in at least one. Optimal predictors
can (in most cases) be based on mixture distributions $\xi$.
Finally we gave an Occam's razor argument that the universal prior
with weights $w_\nu=2^{-K(\nu)}$ is optimal, where $K(\nu)$ is the
Kolmogorov complexity of $\nu$. Of course, optimality always
depends on the setup, the assumptions, and the chosen criteria.
For instance, the universal predictor was not always
Pareto-optimal, but at least for many popular, and for all
decision theoretic performance measures. Bayes predictors are also
not necessarily optimal under worst case criteria \cite{Cesa:01}.
We also derived a bound for the relative entropy between $\xi$ and
$\mu$ in the case of a continuously parameterized family of
environments, which allowed us to generalize the loss bounds to
continuous $\M$.
Furthermore, we discussed the duality between the Bayes-mixture
and expert-mixture (PEA) approaches and results, classification
tasks, games of chances, infinite alphabet, active systems
influencing the environment, and others.

\addcontentsline{toc}{section}{References}
\parskip=0ex plus 1ex minus 1ex

\end{document}